\documentclass[nohyperref]{article}

\usepackage{microtype}
\usepackage{graphicx}
\usepackage{subfigure}
\usepackage{booktabs} % for professional tables

\usepackage{color, colortbl}
\definecolor{LightCyan}{rgb}{0.7529411765,0.9058823529,0.8980392157}

\usepackage{hyperref}
\usepackage{url}

\usepackage[accepted]{icml2022}

\usepackage{amsmath}
\usepackage{amssymb}
\usepackage{mathtools}
\usepackage{amsthm}

\usepackage[capitalize,noabbrev]{cleveref}

\usepackage{multirow}

\usepackage{csquotes}

\theoremstyle{plain}
\newtheorem{theorem}{Theorem}[section]

\newtheorem{hypothesis}[theorem]{Hypothesis}

\theoremstyle{definition}

\theoremstyle{remark}

\usepackage[textsize=tiny]{todonotes}

\newcommand{\potentialTitle}{How Useful are Gradients for OOD Detection Really?}
\newcommand{\Prob}{\mathbb{P}}
\newcommand{\E}{\mathbb{E}}
\newcommand{\norm}[1]{\left\lVert#1\right\rVert}

\newcommand{\R}{\mathbb{R}}

\newcommand{\bx}{\mathbf{x}}
\newcommand{\bh}{\mathbf{h}}
\newcommand{\bp}{\mathbf{p}}
\newcommand{\bxtilde}{\Tilde{\bx}}

\newcommand{\egname}{\textsc{ExGrad}}
\newcommand{\gn}{\textsc{GradNorm}}
\newcommand{\batchgrad}{\textsc{BatchGrad}}

\icmltitlerunning{\potentialTitle}

\begin{document}

\twocolumn[
\icmltitle{\potentialTitle}




\begin{icmlauthorlist}
\icmlauthor{Conor Igoe}{cmu}
\icmlauthor{Youngseog Chung}{cmu}
\icmlauthor{Ian Char}{cmu}
\icmlauthor{Jeff Schneider}{cmu}
\end{icmlauthorlist}

\icmlaffiliation{cmu}{Machine Learning Department, Carnegie Mellon University, Pittsburgh, PA}
\icmlkeywords{Machine Learning}

\vskip 0.3in
]



\printAffiliationsAndNotice{}  % leave blank if no need to mention equal contribution

\begin{abstract}

% One critical challenge in deploying highly performant machine learning models in real-life applications is out of distribution (OOD) detection. Given a predictive model which is accurate on in distribution (ID) data, an OOD detection system will further equip the model with the option to defer prediction when the input is novel and the model has little confidence in prediction. Many previous methods of formulating an OOD detector require additional training of auxiliary models or careful hyperparameter tuning. In this work, we show that an OOD detector can be derived through gradients alone. This method requires no additional training, no hyperparameter tuning, and outperforms current state of the art OOD methods across benchmark experiments. We further do an in-depth analysis on why test time gradients are well suited for OOD detection.

% 1. recent work has described the importance of gradients for OOD
% 2. we demonstrate that the strong performance has been misattributed to gradients, and should instead be attributed to ensembling of learned features and predicted distributions
%     2.1. comparison of various gradient-based OOD techniques
%     2.2 theory analysis of best-performing gradient-based technique
%     2.3 discussion of the inability for gradients to extract feature information from a single test point, and the degradation of performance with the introduction of the ID batch during backpropagation
% 3. we leverage this insight to propose new SotA OOD detectors based on encodings & predicted distributions

One critical challenge in deploying highly performant machine learning models in real-life applications is out of distribution (OOD) detection. Given a predictive model which is accurate on in distribution (ID) data, an OOD detection system will further equip the model with the option to defer prediction when the input is novel and the model has little confidence in prediction.
There has been some recent interest in utilizing the gradient information
in pre-trained models for OOD detection. 
While these methods have shown competitive performance,
there are misconceptions about the true mechanism underlying them,
which conflate their performance with the necessity of gradients.
In this work, we provide an in-depth analysis and comparison of gradient based methods and elucidate the key components that warrant their OOD detection performance.
We further propose a general, non-gradient based method of OOD detection which improves over
previous baselines in both performance and computational efficiency.

% One critical challenge in deploying highly performant machine learning models in real-life applications is out of distribution (OOD) detection. Given a predictive model which is accurate on in distribution (ID) data, an OOD detection system will further equip the model with the option to defer prediction when the input is novel and the model has little confidence in prediction. Many previous methods of formulating an OOD detector require additional training of auxiliary models or careful hyperparameter tuning. On the other hand, recent work in gradient-based methods claim to outperform current state of the art OOD methods across benchmark experiments with no additional training and hyperparameter tuning. We do an in-depth analysis on whether test-time gradients play an essential role for OOD detection.

\end{abstract}

\section{Introduction}

Recent advances in algorithms, models, and training infrastructure
have brought about unprecedented performance of 
machine learning (ML) methods, across a wide range of 
data types and tasks.
Despite their demonstrated potential on benchmark settings and domains, 
one obstacle which limits ML methods' applicability 
in real-world applications is 
the \textit{uncertainty} or \textit{confidence} of the predictions.
Without any deliberate mechanisms, ML models will output 
a prediction for any given input, 
and the question of whether this prediction can be
\textit{trusted} will be especially critical 
in many high-risk decision-making settings 
(e.g. self-driving cars \citep{agarwal2021learning}, 
physical sciences \citep{char2021model, boyer2021machine}, 
and healthcare \citep{zhou2020mortality}).
This risk is further exacerbated with deep learning
where the interpretability of models are often limited
\citep{rudin2019stop}).

It is unrealistic for one to expect to train a model that has perfect predictions for all possible inputs, partly because real-world datasets are limited in their scope. Thus in lieu of trying to make predictions for all test inputs, one can attempt to first detect whether the input is covered by the support of the training data. This is the motivation behind \textit{OOD detection}.

% In lieu of trying to perfect predictions for all possible inputs,
% it can be more efficient to decipher when an input is likely
% to produce an uncertain prediction, which is the main
% goal of OOD detection.

Among the diverse approaches to OOD detection for image recognition,
some works have suggested utilizing the information
in gradients to derive efficient and performant methods for OOD detection 
\citep{liang2017enhancing, lee2020gradients, agarwal2020estimating, huang2021importance}.
% \citet{huang2021importance}, in particular, stands out, 
% both as an efficient method which requires no additional training
% or hyperparameter tuning, and a competitive baseline which
% improves previous SoTA performance. 
% \ysc{should we mention here that
% this is SoTA \textit{only for output-based, post-hoc methods?}}.

%%%%% Dropping Simplex Figure
% \begin{figure}[t]
%     \centering
%     % \includegraphics[width=1\linewidth]{ICML2022/figures/score_fns.pdf}
%     \includegraphics[width=\columnwidth]{ICML2022/figures/score_fns.pdf}
%     \vspace{-5mm}
%     \caption{\textbf{Output Score Components}. 
%     The probability output components (V components from Section \ref{sec:explore_encoding_output}) are shown 
%     on a 3-class probability simplex where each of the 3 vertices signifies probability 1 to a single class.
%     From left to right, these are the output score terms used for \egname, MSP, and \textsc{GradNorm}.
%     Lighter to darker shades indicate lower to higher values.}
%     \label{fig:score_fns}
% \end{figure}
%%%%% Dropping Simplex Figure

We motivate our work by first exploring the key claim that
gradients are useful for OOD detection.
Through a comparison with various extensions of 
gradient-based scores, we analyze the key components 
that actually drive the performance of these methods, 
and we argue that gradient computations are not necessarily 
essential in deriving performant post hoc OOD detectors. 
Rather, these methods ultimately rely on the magnitude of the learned feature embedding and the predicted output distribution.
We thereby refute many of the intuitions that previous works motivate their methods with.
Based on our analysis, we further propose a general 
framework for producing score functions for OOD detection 
and provide a comprehensive empirical evaluation
of various instantiations of the score. This framework improves detection performance across a wide range 
of small and large scale OOD detection benchmarks.

The rest of this paper is structured as follows.
We first provide a formal statement of the problem setting 
and the related works in Section~\ref{sec:prelims}. We then introduce existing and new gradient-based detectors and discuss how both can be simplified into intuitive forms (Section~\ref{sec:grad_method_descriptions}). Following this, we perform empirical evaluations of the methods (Section~\ref{sec:method}) and discuss their implications in Section~\ref{sec:conclusion}.

% In this work, we approach the problem of OOD detection in deep learning
% from the perspective of model gradients.
% A few recent works have suggested the utility of gradient
% information in OOD detection and have proposed methods 
% which are performant on benchmarks.
% However, much of the analysis done in these works are based on
% intuition and not fully principled, which we believe has led to
% a sub-optimal utilization of the gradient information.

% We provide an in-depth discussion into gradient-based methods for OOD detection
% and aim to remedy some the weaknesses in the analysis of existing works.
% Furthermore, we introduce an improved gradient-based algorithm which outperforms the current SoTA method in OOD detection.

% \ysc{
%     Do we wanna also mention the following points in the intro...
%     \begin{itemize}
%         \item point out flaws in analysis done by GradNorm
%         \item GradNorm was motivated by intuition which we debunk, we provide a much more correct intution of \textit{why} gradients are actually useful
%         \item Based on this improved intuition/explanation, provide an even better gradient based method for OOD detection
%         \item Then we do a thorough comparison between GradNorm and EpiGrad
%     \end{itemize}
% }

\begin{figure*}[t]
    \centering
    \includegraphics[width=0.7\linewidth]{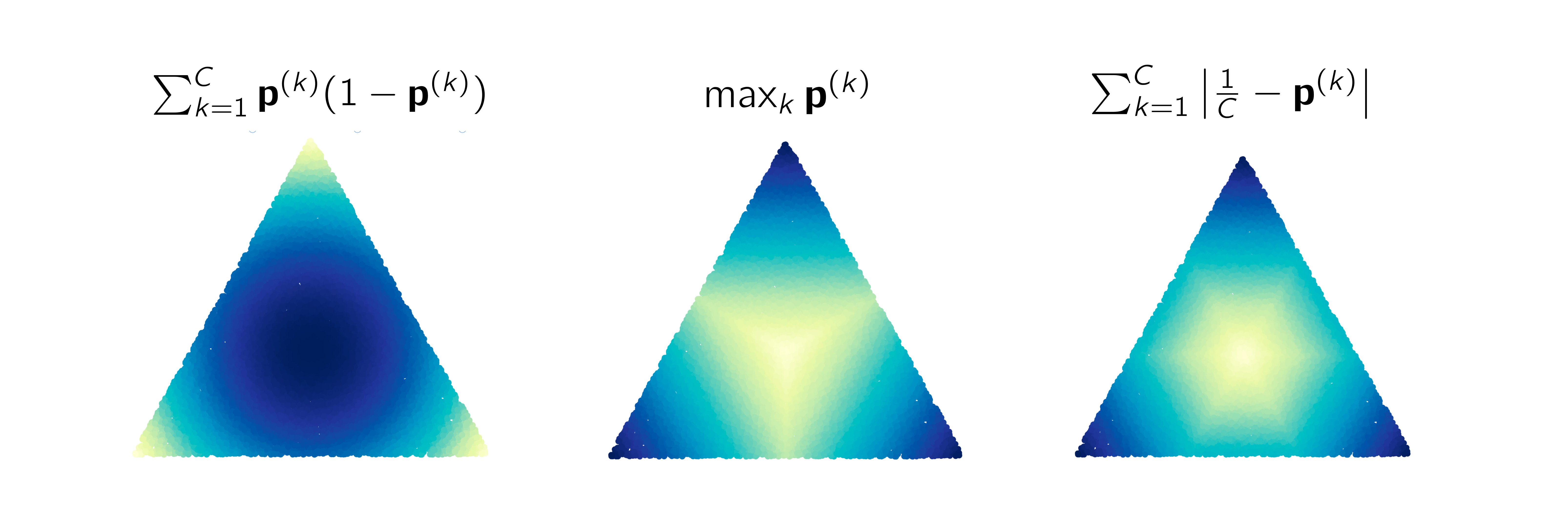}
    \vspace{-5mm}
    \caption{\textbf{Output Score Components} 
    The probability output components ($V$ components from Section \ref{sec:explore_encoding_output}) are shown 
    on a 3-class probability simplex where each of the 3 vertices signifies probability 1 to a single class.
    From left to right, these are the output score terms used for {\egname}, MSP, and {\gn}.
    Lighter to darker shades indicate lower to higher values.}
    \label{fig:score_fns}
\end{figure*}

\section{Preliminaries and Related Work} \label{sec:prelims}

% General set up and problem notation.
\textbf{Problem Setting and Notation}
We focus on the classification setting where the task is to predict a class label $y \in [C]$ given an input $\bx \in \R^{d}$, where $C$ is the total number of possible classes and $d$ is the dimensionality of the input. We assume we have access to training data $\{(\bx_i, y_i)\}_{i=1}^N$ in order to train the parameters $\theta$ of a deep neural network $f_\theta: \R^d \rightarrow \R^C$.
Here, $\theta$ is a collection of all of the weights from each layer
in the network, thus, $\theta = \{W_{l} : l \in [L]\}$, where $l$ denotes
the index of each layer, and we assume the network has $L$ layers in total.

Given an input $\bx$, the network predicts 
a probability vector, $\bp$, 
via the softmax function of the network outputs: 
% $\bp^{(k)}(\bx) = \Prob_{\theta}(Y = k | \bx) = \frac{\textrm{exp}\left(f^{(k)}_\theta(\bx) / T \right)}{\sum_{k'=1}^C\textrm{exp}\left(f^{(k')}_\theta(\bx) / T \right)}$
$\bp^{(k)}(\bx) = \Prob(Y = k | \bx) = \frac{\textrm{exp}\left(f^{(k)}_\theta(\bx) / T \right)}{\sum_{k'=1}^C\textrm{exp}\left(f^{(k')}_\theta(\bx) / T \right)}$
, where the superscripts denote the index of the vector and $T$ is the temperature. If not otherwise specified, we will assume the temperature $T=1$. Although $\textbf{p}(\textbf{x})$ depends on both $\theta$ and $\textbf{x}$ , we will always exclude the former and often exclude the latter when it is clear from context or unimportant. Lastly, we often abuse notation and use $Y \sim \textbf{p}(\textbf{x})$ to mean $Y$ is sampled from the categorical distribution parameterized by $\textbf{p}(\textbf{x})$.

% It is assumed the standard training procedure of minimizing the cross entropy loss w.r.t. a one-hot vector label for the true class, i.e. $\mathcal{L}\left(\bp(\bx), y \right) = - \log \bp^{(y)}(\bx)$.

% % General set up and problem notation.
% \textbf{Image Classification.} In this paper, we focus on the setting of image classification using deep neural notworks. That is, the task is to predict a class label $y \in \{1, 2, \ldots, C\}$ of given an image $\bx \in \R^{d}$, where $C$ are the total number of possible classes and $d$ is the dimensionality of the image. We assume we have access to training data $\{(\bx_i, y_i)\}_{i=1}^N$ in order to train a deep neural network $f_\theta: \R^d \rightarrow \R^C$, where $\theta$ is the parameters of the network. Given an input image $\bx$, we can form a vector of probabilities, $\bp$, where $\bp^{(k)}(\bx) = \Prob(y = k) = \frac{\textrm{exp}\left(f^{(k)}_\theta(\bx)\right)}{\sum_{k'=1}^C\textrm{exp}\left(f^{(k')}_\theta(\bx)\right)}$ and the superscripts denote the index of the vector \ianx{run on fix later, also no temperature is that OK?}. We also assume that the network is trained using cross entropy loss, i.e. $\mathcal{L}\left(\bp(\textbf{x)}), y \right) = - \log \bp^{(y)}(\bx)$.

% What is out of distribution detection here? How do we denote it? What are some examples.
\textbf{The OOD Detection Problem} 
In a real-world scenario, the model may be given an input, $\Tilde{\bx} \in \R^d$, during deployment that is substantially different from any of the 
datapoints in the training set. For example, a classifier trained
to identify numeric digits may be given an image of a cat. 
Since the model's prediction $\bp(\bxtilde)$ cannot be trusted, 
it would be advantageous to flag such instances and defer prediction. This is the problem that OOD detection addresses. 

More formally, the goal in OOD detection is to derive a binary classifier 
which labels whether a given input, $\bxtilde$, is ID (in distribution) or OOD.
This goal is commonly addressed by learning 
a score function $\mathcal{S}: \R^{d} \rightarrow \mathbb{R}$
which quantifies the degree to which the input is OOD.
Existing works have approached this problem of learning the mapping $\mathcal{S}$ from 
various perspectives, which generally vary by \textit{where} the signal to generate $\mathcal{S}$ is extracted from.
Here, we introduce broad groupings of methodologies to provide context for 
our work, and we refer the reader to \citet{yang2021generalized, salehi2021unified}
for an in-depth survey of the field.
% \todo{make this a footnote}

One class of methods focuses on the input space 
and learns score functions based on characteristics
that can be derived from the input features.
For example, distance-based methods are based on 
the intuition that 
OOD data should lie ``far away'' from ID data 
and define $\mathcal{S}$ with distances
between the input point and reference points that 
are representative of ID \citep{lee2018simple, techapanurak2020hyperparameter, van2020uncertainty}.
Meanwhile, density-based methods utilize probabilistic models
to describe the density of ID data and argue that 
OOD points should occur in areas of low densities. 
Thus generative models are often used and the score function is often
derived with the predicted likelihood of 
the test input point \citep{ren2019likelihood, serra2019input, zisselman2020deep}.

Another class of methods aims to directly influence
a predictive model's behavior to OOD data through explicit training. These methods often assume 
access to OOD examples during training and incorporates them into 
a predictive model's training procedure to maximize the separability between ID and OOD inputs. 
This is usually achieved by setting aside actual samples
from an OOD test distribution \citep{hendrycks2018deep, Yu_2019_ICCV, liu2020energy}, or if they are not available, 
by synthesizing OOD examples via adversarial training, perturbations, 
or sampling from boundaries or low density regions \citep{lakshminarayanan2016simple, lee2017training, vernekar2019out}.

Many works have instead turned attention to the information
that can be extracted from a predictive model that is fully 
trained on ID data. 
With the intuition that an ideal predicted distribution should 
have low uncertainty (heavily concentrated on the predicted class)
for an ID point and have high uncertainty (close to a uniform distribution)
for an OOD input, some methods focus on generating scores
with the predicted distribution of pre-trained models \cite{hendrycks2016baseline, linmans2020efficient, liu2020energy}. 
Meanwhile, some works have examined the information available 
when backpropagating gradients of a loss function.
\citet{liang2017enhancing, agarwal2020estimating} utilize information in the gradient w.r.t. the input,
while \citet{lee2020gradients, huang2021importance} examine gradients w.r.t. the model parameters. 
\citet{huang2021importance}, in particular, proposed {\gn}, an efficient post hoc score function which simply
measures the norm of model parameter gradients and thus requires no additional training or hyperparameter tuning. Despite this, {\gn} achieves State-of-the-Art (SotA) performance among post hoc methods. We explain {\gn} in more detail in the next section.

% By improving upon previous State-of-the-Art (SotA) performance across benchmark experiments, {\gn} claimed the high utility of model
% gradients for OOD detection.

% Drawing motivation from its efficiency and performance,
% in the next section, we further investigate the claim that
% gradients are useful by analyzing {\gn} and other gradient-based scores.
% % We further explore this claim in the next section by 
% % revisiting \textit{GradNorm} and deriving other gradient-based scores.
% \ysc{I def need to motivate this more...}

% As a naive example \ianx{better example?} of this is a model trained to recognize hand-written digits zero through eight being given the digit nine.

% % Definition of an OOD detector.
% Of course, one would like to know when these instance occur since it is unlikely that the model will classify $\Tilde{\bx}$ correctly. To combat this, one can derive an OOD detector, $h: \R^d \rightarrow \{0, 1\}$, that takes an image and outputs $0$ if the image is in distribution and $1$ if the image is out of distribution.

% \input{related}

\section{Gradient Based OOD Methods}\label{sec:grad_method_descriptions}

% There are gradient methods... \ianx{YOUNG TO DO HERE}
% \textbf{GradNorm \cite{huang2021importance}}
\citet{lee2020gradients} initially proposed using model gradients as a signal
to detect OOD data. Specifically, they provide a fully trained network with an input point and backpropagate the cross entropy loss between a uniform probability distribution and the network's predicted class probabilities. We summarize their intuition for this algorithm in the following hypothesis:

\begin{hypothesis}\label{hyp:feature_extraction}
\textbf{The Feature-Extraction Hypothesis} If learning a point $(\tilde{\mathbf{x}},\tilde{y})$ requires a large change in a well-trained network, then $\tilde{\mathbf{x}}$ must have been novel, because most of a deep network is dedicated to feature extraction, which by assumption should already perform well on ID data.
\end{hypothesis}

% Their hypothesis is that, an OOD input will require a large change in the network
% to learn new features representations and associations to the label.

% \textbf{Feature-Extraction Hypothesis.}
% As described in the introduction, to the best of the authors' knowledge, the OOD literature has so far not presented a compelling theory that illuminates the unique role of gradients in achieving SotA OOD detection performance. One preliminary theory has been described in a number of papers \cite{lee2020gradients, huang2021importance}, which we refer to as the \textbf{\textit{feature-extraction hypothesis}}. We summarize this hypothesis as follows: \textit{if learning a point $(\tilde{\mathbf{x}},\tilde{y})$ requires a large change in a well-trained network, then $\tilde{\mathbf{x}}$ must have been novel, because most of a deep network is dedicated to feature extraction, which by assumption should already perform well on ID data.} In the following section, we perform experiments that challenge this hypothesis, showing the inability of gradients to measure feature extraction.

Meanwhile, {\gn} contends the \textit{opposite case}: they argue that
trying to fit the predicted distribution of \textit{ID} data to a uniform distribution label will necessitate higher gradients to the model parameters than with OOD data. 
We note that this intuition relies on the predicted
distributions actually displaying high entropy for OOD inputs and low entropy for ID inputs. 
% the entropy of the predicted distribution, which many works have actually dissuaded from relying on \citep{}.

Despite conflicting intuitions, both works ultimately propose taking 
the KL divergence (identically, the cross-entropy loss) w.r.t. the uniform distribution and measuring a norm of the model gradients as a score function.
Through extensive empirical ablations, {\gn} hones in on the $L_1$ norm of the last layer gradients of the KL-divergence w.r.t. the uniform distribution as the score function,
\begin{align}
    \mathcal{S}_{GN}(\bx) = 
    \norm{ 
        \frac{\partial D_{KL}(\mathbf{u} || \text{softmax}(f_\theta (\bx))}{\partial W_L}
        }_{1}.
\end{align}
We can expand the divergence term and simplify this score into the following expression:
\begin{align}\label{eq:gradnorm}
    \mathcal{S}_{GN}(\bx) = \norm{\E_{Y \sim \text{unif}}\left[\nabla_{W_L}\log \textbf{p}^{(Y)} \right]    }_{1}
\end{align}
That is, {\gn} measures the norm of the
expected gradient according to a uniform labeling distribution.

% \textbf{Alternative Gradient Approaches.}
\textbf{Another Gradient Approach}
% If, indeed, by any intuition or reasoning, 
% model gradients do provide a useful signal, we note that Eq.~\ref{eq:gradnorm} is not necessarily a unique measure.
We note that Eq.\ref{eq:gradnorm} is not a unique measures that leverages gradients, and we propose our own novel score:
\begin{align}\label{eq:epigrad}
    \mathcal{S}_{EG}(\textbf{x}) = \E_{Y \sim \textbf{p}(\textbf{x})}\left[\norm{\nabla_\theta\log \textbf{p}^{(Y)} }_{p} \right]
\end{align}
This is a natural score since it simply measures the average
norm of the gradients according to the model's predicted distribution.

Due to the outer positioning of the expectation, we refer to the score function in Eq. \ref{eq:epigrad} as {\egname}. Two key differences in this score are 1) the label distribution of $Y$ comes from the model's own predicted distribution, $\bp$, not the uniform distribution, and 2) this calculates the expected norm of the gradient,
whereas Eq. \ref{eq:gradnorm} calculates the norm of the expected gradient.

While the effect from these changes may seem unclear at first,
the decomposition of the scores in the following subsection elucidates what calculations are actually being done.

Other gradient-based scores are readily plausible, and we visit a suite of scores with empirical comparisons in Section~\ref{sec:different_grads}.

\subsection{Decomposition of Gradient Methods}\label{sec:decomposition}
To give further grounds for their method, 
\citet{huang2021importance} analyze Eq.~\ref{eq:gradnorm} and show that it can be decomposed into two terms: one that is characterized by the size of the magnitude of the encoding fed to the last layer of the network, and one that is characterized by the output of the network. In particular, they derive
\begin{align} \label{eq:gradnorm_decomp}
    \mathcal{S}_{GN}(\bx) = \frac{1}{T}\norm{\textbf{h}}_1 \sum_{k=1}^C \left| \frac{1}{C} - \textbf{p}^{(k)} \right|
\end{align}
where $\textbf{h}$ is the encoding being fed to the last layer of the network. Following their notation, we will denote $U$ to be the part of the score characterized by $\textbf{h}$ and $V$ to be the part characterized by the network output, i.e. here $U = \norm{\textbf{h}}_1$, $V = \frac{1}{2}\sum_{k=1}^C \left| \frac{1}{C} - \textbf{p}^{(k)} \right|$, and $\mathcal{S}_{GN} = \frac{2}{T} U V$. \citet{huang2021importance} show that $U$ and $V$ have the same trend where their values are large for ID points and lower for OOD points. While each can be used as OOD detectors alone, the product of the two terms results in stronger performance.

Although not mentioned in their analysis, the $V$ term in {\gn}'s score is simply the total variation (TV) distance between a discrete uniform distribution and the model's predicted distribution. This characterization of $V$ gives insight as to why {\gn} works. When the input image is in distribution, the network will likely have higher confidence, making the TV distance between $\textbf{p}$ and the discrete uniform large. 

\textbf{Decomposition of {\egname}} Doing a similar analysis, our alternative gradient-based detector, {\egname}, can be broken down in a similar way. In particular,
\begin{align*}
    \mathcal{S}_{EG}(\bx) = \frac{2}{T} U V 
\end{align*}
\vspace{-3mm}
where
\begin{align} \label{eq:epigrad_decomp}
    U = \norm{\textbf{h}}_1 \hspace{5mm} V = \sum_{k=1}^C \textbf{p}^{(k)} (1 - \textbf{p}^{(k)}) 
\end{align}
The full derivation is shown in Appendix~\ref{app:epi_derivation}. 
Like with {\gn}, the $V$ term for {\egname} turns out to be an interpretable quantity. Let $B_k \sim \textrm{Bernoulli}(\textbf{p}^{(k)})$ be the random variable corresponding to the event that $\bx$ belongs to class $k$. Then, $V = \sum_{k=1}^C \textrm{Var}(B_k)$. Intuitively, when the input image is in distribution and there is high confidence on a single class, the variance of each Bernoulli random variable will be low. Note that this is the opposite trend of the TV distance and $\norm{\textbf{h}}_1$. A visual comparison of these scores can be seen in Figure~\ref{fig:score_fns}.

This general $UV$-style score will be of particular interest to us throughout this paper, and we will often refer to it as an ``Encoding-Output'' composition as it relies on both the networks encoding for the image (at least at the penultimate layer) and its output. 

\section{Empirical Evaluation} \label{sec:method}

\begin{table*}[ht]
\renewcommand{\arraystretch}{1.5}
\centering
\begin{tabular}{|lcc|}
\hline
                        \multicolumn{1}{|c}{\textbf{Score Expression}} & \multicolumn{1}{c}{\textbf{AUROC}} & \multicolumn{1}{c|}{\textbf{Gradient Depth}}\\
                        \hline
$\lVert \mathbb{E}_{Y\sim \mathbf{p}}\left[ \nabla_\theta\log\mathbf{p}^{(Y)} \right] \rVert_2^2$       & 0.725 ($\pm$ 0.086)                                         & Deep                   \\
$\lVert \mathbb{E}_{Y\sim \mathbf{p}}\left[ \nabla_\theta\log{\mathbf{p}^{(Y)}} \right] \rVert_2^2$    & 0.741 ($\pm$ 0.081)                                 & Shallow                \\
$\mathbb{E}_{Y\sim \text{Uniform}}\left[\lVert\nabla_\theta\log { \mathbf{p}^{(Y)}} \rVert_1 \right]$      & 0.825  ($\pm$ 0.120)                                   & Deep                           \\
$\mathbb{E}_{Y\sim \text{Uniform}}\left[\lVert\nabla_\theta\log { \mathbf{p}^{(Y)}} \rVert_1 \right]$    & 0.850     ($\pm$ 0.135)                                  & Shallow                           \\
$\mathbb{E}_{Y\sim \text{Uniform}}\left[\lVert\nabla_\theta\log { \mathbf{p}^{(Y)}} \rVert_2^2 \right]$     & 0.867   ($\pm$ 0.148)                                    & Deep                            \\
$\mathbb{E}_{Y\sim \text{Uniform}}\left[\lVert\nabla_\theta\log { \mathbf{p}^{(Y)}} \rVert_2^2 \right]$       & 0.887     ($\pm$ 0.107)                                  & Shallow                            \\
% \textbf{Deep EGB}       & 0.787                                 & 0.224                                     \\
$\left\lVert \nabla_\theta \mathbb{E}_{Y\sim \text{Uniform}}\left[\log{\mathbf{p}^{(Y)}}\right]\right\rVert_1$  \quad ({\gn})      & 0.892   ($\pm$ 0.087)                                                           & Deep  \\
$\left\lVert \nabla_\theta \mathbb{E}_{Y\sim \text{Uniform}}\left[\log{\mathbf{p}^{(Y)}}\right]\right\rVert_1$  \quad ({\gn})   & 0.906    ($\pm$ 0.092)                                      & Shallow                  \\
$\mathbb{E}_{Y\sim \mathbf{p}} \left[ \frac{\log \mathbf{p}^{(Y)}}{\mathbf{p}^{(Y)}} \lVert\nabla_\theta\log\mathbf{p}^{(Y)} \rVert_2^2 \right]$ & 0.910    ($\pm$ 0.090)                                  & Shallow               \\
$\mathbb{E}_{Y\sim \mathbf{p}}\left[\lVert\nabla_\theta\log { \mathbf{p}^{(Y)}} \rVert_2^2 \right]$  & 0.919         ($\pm$ 0.041)                                            & Shallow           \\
$\mathbb{E}_{Y\sim \mathbf{p}}\left[\lVert\nabla_\theta\log { \mathbf{p}^{(Y)}} \rVert_2^2 \right]$      & 0.921        ($\pm$ 0.034)                                   & Deep                    \\
$\mathbb{E}_{Y\sim \mathbf{p}} \left[ \frac{\log \mathbf{p}^{(Y)}}{\mathbf{p}^{(Y)}} \lVert\nabla_\theta\log\mathbf{p}^{(Y)} \rVert_2^2 \right]$    & 0.921         ($\pm$ 0.106)                            & Deep                   \\
$\mathbb{E}_{Y\sim \mathbf{p}}\left[\lVert\nabla_\theta\log { \mathbf{p}^{(Y)}} \rVert_1 \right]$ \quad ({\egname})  & 0.925      ($\pm$ 0.047)                                    & Shallow                     \\
% \textbf{Shallow EGB}    & 0.925                                 & 0.044                                     \\
$\mathbb{E}_{Y\sim \mathbf{p}}\left[\lVert\nabla_\theta\log { \mathbf{p}^{(Y)}} \rVert_1 \right]$  \quad ({\egname})   & 0.926      ($\pm$ 0.053)                                   & Deep                 \\
\rowcolor{LightCyan} $\sum_{k=1}^C \textbf{p}^{(k)} (1 - \textbf{p}^{(k)}) $    \quad ({\egname} $V$ term)    & 0.933         ($\pm$ 0.063)                                      & \textit{N/A} \\
\rowcolor{LightCyan} $\sum_{k=1}^C \left| \frac{1}{C} - \textbf{p}^{(k)} \right|$  \quad ({\gn} $V$ term)       & 0.935      ($\pm$ 0.064)                            &        \textit{N/A}                      \\
\hline
\end{tabular}
    \caption{\textbf{AUROC for Gradient-Based OOD Detectors on Small-Scale Experiments} The leftmost column shows the definition of each score function. The Gradient Depth column signifies whether gradients were taken with respect to all parameters (``Deep'') or just the last layer's parameters (``Shallow''). For each row, the mean AUROC is reported with standard deviations in parentheses, both calculated across the 6 ID-OOD dataset combinations from \{MNIST, CIFAR-10, SVHN\}. Rows are sorted by mean AUROC. Note that the last two rows are the highest performing score functions and correspond to scores that do not involve any gradient calculations.}
    \label{tab:small_scale}
\end{table*}

% \ysc{
% Roadmap for presenting experiment results
% \begin{itemize}
%     \item First aim to answer, compare performance of different forms of gradient based scores
%     \item To what extent is good performance due to gradients and to what extent is it due to encoding-output composition?
%     \item Does the feature extraction hypothesis explain gradient based OOD detection performance?
% \end{itemize}

% First evaluate on small-scale benchmarks how each of the scores compare.

% We notice that a simple output-based score achieves competitive results.

% 4.1 should just be the variants of the score functions

% Table 2 --  same set of results for both small and large scale
% }
The previous section introduced variations to 
existing gradient-based scores and decomposed
the scores into interpretable components. 
This naturally begs the questions, which scores 
actually perform well, and what should their favorable performance be attributed to?
In this section, we explore these questions
by comparing the performance of
multiple gradient-based scores against simpler, non-gradient based approaches
on OOD detection tasks in image classification.
Specifically, the goals of these experiments are as follows:
% In this section, we describe the results of a small-scale image classification experiment comparing multiple gradient-based OOD detection methods against simpler, non-gradient based approaches.
% The goal of these experiments is threefold: 
\begin{enumerate}
    \item to investigate whether different gradient-based scores are useful for OOD detection or whether only particular variants perform well
    (i.e. is it essential that the gradient-based detector is derived by taking the derivative of the KL divergence?);  
    % (i.e. is backproping KL divergence some sort of magical catch all method?);
    \item to examine the plausibility of the feature-extraction hypothesis in explaining the strong performance of gradient-based OOD detection methods;
    % (i.e. do we really buy the story that's been spun so far about gradients?);
    \item to investigate the claim that gradient-based approaches offer unique performance advantages in the context of post hoc OOD detection tasks.
    % (i.e. do we really need gradients to achieve SotA post-hoc OOD detection performance?)
    % \todo{We better make sure the scope is this from the beginning}  
\end{enumerate}
 Indeed, we show that SotA OOD detection performance is achievable by leveraging information solely from the predicted class distribution and latent encoding, calling into question the additional utility gained from gradient based score functions. Furthermore, we find that there is significant variability in performance across tasks for different gradient-based score functions. Moreover, we show that gradient-based methods
are often worse than
% \textit{dominated} by
computationally simpler approaches that require no backpropagation. 
% These simpler score functions are inspired by analysis in \citet{huang2021importance} and Section 3.1.
Lastly, we perform experiments that challenge the plausibility of one emerging theory that attempts to explain the role of gradients in SotA OOD methods.

\subsection{Investigating the Performance of Gradient-Based Score Variants} \label{sec:different_grads}
\textbf{Experimental Set Up} 
% Here we describe our first experimental setup and the various OOD methods, the results of which are described in Table \ref{tab:small_scale}.
A key motivation of our work is in the design of post hoc OOD detection mechanisms. In light of this, for each OOD method, we take a deep neural network pre-trained on a particular dataset (the ID dataset) and use data from one of the remaining datasets as OOD data. These pre-trained networks were obtained from a popular open-source library \footnote{\url{https://github.com/aaron-xichen/pytorch-playground}} of pre-trained neural network image classifier models and achieve competitive performance on ID data.

We investigate the performance of seven gradient-based OOD scores, one of which has been previously described in the literature ({\gn}) and another which we described in detail in Section 3 ({\egname}). We include additional experimental results on natural variants that can be derived by simple design choices in implementation, specifically by interchanging norms and expectations, choice of norm and choice of distribution to generate the synthetic label at test time. Finally, we compare these gradient-based methods against two non-gradient based approaches inspired by \citet{huang2021importance} and our analysis in Section 3. We refer to these scores as ``$V$ term'' scores.
% We refer to these scores as ``$2^{\text{nd}}$ term'' scores given that their inspiration is from the second terms appearing in their respective decomposition analyses. 

We used MNIST \cite{deng2012mnist}, SVHN \cite{Netzer2011ReadingDI} and CIFAR-10 \cite{Krizhevsky09learningmultiple} as our base datasets for the first experimental setup, and use 10,000 samples from the test split of each dataset to form our ID and OOD datasets. Each OOD method defines a score function $\mathcal{S}({\mathbf{\tilde{x}}})$ which we then use to define an OOD classifier ${\mathbf{I}[ \mathcal{S}({\mathbf{\tilde{x}}}) > \epsilon ]}$. We then calculate AUROCs for each method, varying the threshold $\epsilon$. Table \ref{tab:small_scale} describes the mean and standard deviation of the AUROC for each method across the six ID-OOD dataset combinations. Full experimental results showing performance for each ID-OOD combination are available in Appendix \ref{app:tables}.

The columns ``Deep'' and ``Shallow'' refer to variants of each method that perform gradients w.r.t. all parameters of the network (Deep) or w.r.t. just the parameters of the final layer (Shallow), i.e. the weights of the layer that generate logits. 

\textbf{Analysis of Results} We first observe that, on average over the 6 ID-OOD benchmark tasks, {\egname} outperforms the previous SotA gradient-based post hoc OOD method, {\gn}. Moreover, \textit{no method consistently dominates all other methods}, suggesting that the singular focus in the literature on backpropagating KL divergence losses is unwarranted.

Our next observation is that, in instances where extending gradients to the entire network produces change in performance, such changes are at best modest gains, and can in fact \textit{hurt} performance. This aligns with results found in \citet{huang2021importance} and serves as further evidence against the feature-extraction hypothesis.

Lastly, we note that the highest performing score functions are the \textit{non-gradient based} approaches. In particular, the two score functions inspired from the decomposition analyses improve beyond their best gradient-based variants by 2.9 and 0.7 percentage points respectively. 

These observations provide the basis for a general framework to achieve high-performing post hoc OOD detectors that do not rely on calculating test-time gradients: combining a norm of learned encodings with a function of the predicted class distribution. In the following section, we provide additional experimental details that leverage this proposed template to further illustrate the ability to achieve high performance without relying on test-time gradients.

\begin{table*}[t]
\renewcommand{\arraystretch}{1.01}
    \centering
    \begin{tabular}{|cc|cccc|ccc|cc|}
        \hline
          \multicolumn{2}{|c|}{Dataset} & \multicolumn{4}{c|}{$V$ Only} & \multicolumn{3}{c|}{$\norm{\textbf{h}}_1 V$} & \multicolumn{2}{c|}{Gradient-Based} \\
         \textbf{ID} & \textbf{OOD}  &   \textbf{MSP} &   \textbf{Energy} &   \textbf{VarSum} &    \textbf{TV} &   \textbf{MSP} &   \textbf{Energy} &   \textbf{VarSum} &   \textbf{GradNorm} &   \textbf{ExGrad} \\
        \hline
         \multirow{2}{*}{MNIST}         & CIFAR10     & 0.971 &    0.957 & 0.972       & \textbf{0.974} & 0.907 & 0.928       &    0.867 & 0.896       & 0.963       \\
                  & SVHN        & 0.987 &    0.989 & 0.988       & \textbf{0.990} & 0.978 & 0.980       &    0.949 & 0.971       & 0.975       \\ \hline
         \multirow{2}{*}{CIFAR10}       & MNIST       & 0.924 &    0.831 & 0.925       & \textbf{0.928} & 0.913 & 0.860       &    0.895 & 0.913       & 0.921       \\
               & SVHN        & 0.943 &    0.926 & 0.944       & 0.948       & 0.994 & 0.957       &    0.995 & \textbf{0.995} & 0.919       \\ \hline
         \multirow{2}{*}{SVHN}           & MNIST       & 0.813 &    0.807 & 0.813       & 0.811       & 0.748 & 0.740       &    0.637 & 0.735       & \textbf{0.840} \\
                   & CIFAR10     & 0.955 &    0.953 & \textbf{0.956} & 0.955       & 0.936 & 0.914       &    0.844 & 0.926       & 0.931       \\ \hline
         \multicolumn{2}{|c|}{Average}     & 0.932 &    0.91  & 0.933       & \textbf{0.935} & 0.913 & 0.896       &    0.865 & 0.906       & 0.925       \\ \hline \hline
         \multirow{4}{*}{ImageNet}      & iNaturalist & 0.876 &    0.885 & 0.884       & 0.885       & 0.892 & \textbf{0.917} &    0.748 & 0.904       & 0.769       \\
               & SUN         & 0.782 &    0.852 & 0.790       & 0.830       & 0.818 & \textbf{0.910} &    0.787 & 0.890       & 0.666       \\
               & Places      & 0.767 &    0.813 & 0.771       & 0.789       & 0.795 & \textbf{0.872} &    0.729 & 0.849       & 0.689       \\
               & Textures    & 0.744 &    0.758 & 0.750       & 0.761       & 0.778 & \textbf{0.817} &    0.729 & 0.811       & 0.651       \\ \hline
         \multicolumn{2}{|c|}{Average}    & 0.792 &    0.827 & 0.799       & 0.816       & 0.821 & \textbf{0.879}       &    0.748 & 0.864       & 0.694       \\ \hline
    \end{tabular}
    \caption{\textbf{AUROC scores} The table shows AUROC scores for several detectors grouped by category. The leftmost group uses only the output of the network ($V$ only), the second is the product of these measures with the 1-norm of the encoding passed to the last layer of the network $\norm{\textbf{h}}_1 V$, and the last grouping is detectors that leverage gradients of the last layer of the network. The highest AUROC found for each OOD task is bolded. We separately report the average of the ImageNet baselines and all of the other baselines. Note that we do show $\norm{\textbf{h}}_1 TV$ in this table since it is exactly \textsc{GradNorm}.}
    \label{tab:all_aucs}
\end{table*}

\subsection{Exploring Encoding-Output Compositions}
\label{sec:explore_encoding_output}
% I took out "Alternative" from
\textbf{The ImageNet Benchmark} In addition to the previous experimental set up, in this section we also evaluate on the large-scale ImageNet benchmark proposed by \citet{huang2021mos}. For this benchmark, the ImageNet-1k dataset \citep{deng2009imagenet} is used for the ID dataset. This ID dataset is different from MNIST, CIFAR10, and SVHN in that it is composed of higher resolution images and has $C=$ 1,000 classes instead of $C = 10$. The iNaturalist \citep{van2017inaturalist}, SUN \citep{xiao2010sun}, Places \citep{zhou2017places}, and Textures \citep{cimpoi2014describing} datasets are used as OOD datasets. This benchmark uses the 50,000 images in the ImageNet validation set as the ID data and 10,000 images for each of the OOD datasets (except for Textures which uses 5,640). The pre-trained model is from Google BiT-S\footnote{\url{https://github.com/google-research/big_transfer}} \citep{kolesnikov2020big} and uses the ResNetv2-101 architecture \citep{he2016identity}. Our code for these experiments was built on top of the code from \citet{huang2021importance}\footnote{\url{https://github.com/deeplearning-wisc/gradnorm_ood}}, and more details about this baseline can be found in their paper.

% Inspired by the results above, here we investigate whether scores that are based on encoding-output compositions alone, and not derived from gradients, are competitive.

\begin{figure*}[ht]
    \centering
    \includegraphics[width=1\linewidth]{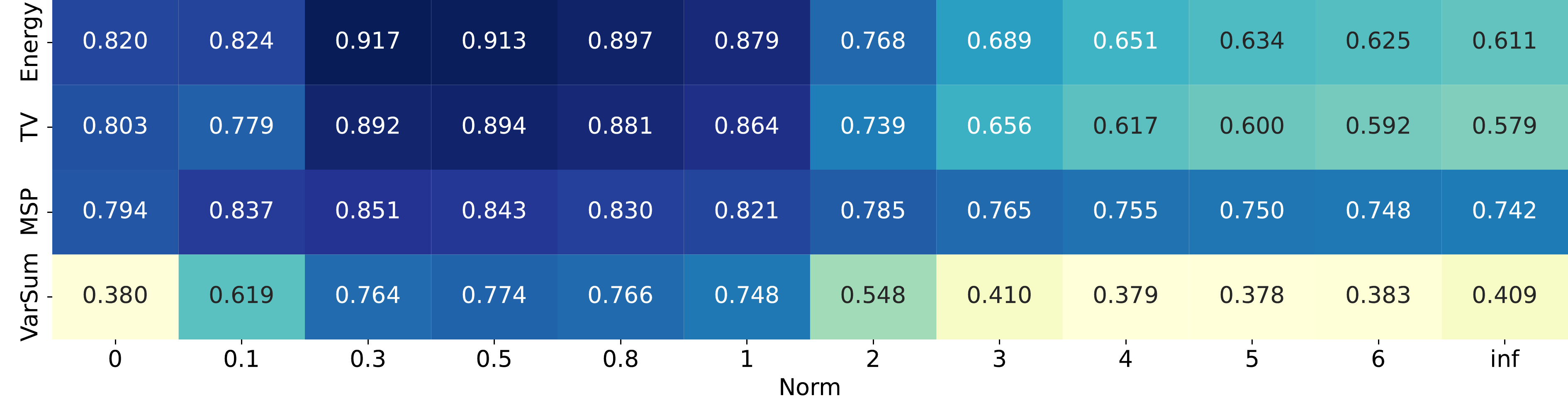}
    \vspace{-5mm}
    \caption{\textbf{Average AUROC over ImageNet Benchmark} Each cell of the heatmap shows the average AUROC for a different configuration of probability output score (y-axis) and order of the norm on the encoding fed to the last layer (x-axis).}
    \label{fig:norm_variation}
\end{figure*}

\textbf{Experimenting with the Encoding-Output Composition} 
In what follows, we test a variety of detectors that adhere to the encoding-output composition described in Section~\ref{sec:decomposition}. In particular, each of the methods we test here are a product of a $U$ term, which is a function of the encoding $\textbf{h}$, and a $V$ term, which is a function of the outputted probability vector $\textbf{p}$. We test all combinations of $U$ $\in \{1, \norm{\bh}_{1}\}$ and $V$ $\in$ \{Energy, VarSum, MSP, TV\}. 

Here, $U = 1$ denotes not using any encoding of the input and only using the output score (i.e. $V$ term).
For the $V$ term, TV is the TV distance between $\textbf{p}$ and a discrete uniform distribution (V term from Eq. \ref{eq:gradnorm_decomp}). The scores for Energy \citep{liu2020energy} and MSP \citep{hendrycks2016baseline} are as follows:
\begin{align*}
    \mathcal{S}_{\textrm{Energy}} &= T \log \sum_{k=1}^C e^{f^{(k)}(\textbf{x})/T} \\
    \mathcal{S}_{\textrm{MSP}} &= \max_{k \in [C]} \textbf{p}^{(k)}
\end{align*}
Note that the energy score we use here is the negative version of the one originally introduced in \citet{liu2020energy}. Like before, we assume that $T=1$ for all experiments.

VarSum is a term inspired by the decomposition of {\egname} (i.e the $V$ term from Eq. \ref{eq:epigrad_decomp}). Because $V = \sum_{k = 1}^C \textbf{p}^{(k)} (1 - \textbf{p}^{(k)})$ is anti-correlated with every other score, we make VarSum a correlated version of this. In particular, $\textrm{VarSum} = 1 - \sum_{k = 1}^C \textbf{p}^{(k)} (1 - \textbf{p}^{(k)})$.
Each of the $V$ terms explored here are visualized 
in Figure~\ref{fig:score_fns}, with the exception of Energy, since logits cannot be deduced from probabilities alone. 

Table~\ref{tab:all_aucs} displays the OOD detection performance 
for each of these methods in both the small-scale benchmark setting (with MNIST, CIFAR10, SVHN datasets) 
and the large-scale benchmark setting with the
ImageNet dataset. 
The small-scale experiments assume the same setting as Section~\ref{sec:different_grads}. 

% Deprecated
% For the large-scale setting, 
% we assume the same setting as \cite{huang2021importance} where we take ImageNet\todo{1k} dataset as ID, assume a Resnetv2-101 model pretrained on ImageNet1k, test the OOD detection performance across 5 different datasets \{iNaturalist, SUN\}, and report the mean AUROC and FPR\textsubscript{95} across the 5 datasets. 
% The last column, ``Average'' indicates the averaged performance across all datasets in both the small and large scale settings.

\textbf{Analysis of Results}
It is immediately clear that there is a significant difference between the ImageNet baseline and the other baselines, and as such, we average the AUROCs separately. Starting with the small-scale setting, we find the best performers to be methods which look at the outputted predicted distribution only. Interestingly, the two best scores appear to be VarSum and the TV distance, which to the best of our knowledge, have not been recommended for OOD detection by previous works. 

Literature generally warns against exclusively using the model's predicted distribution for OOD detection \citep{Hein_2019_CVPR, kirsch2021pitfall}
since even OOD points can produce confident predictions with probabilities that are highly concentrated on a single class. 
These results indicate that this issue is milder in 
smaller scale models (MNIST, SVHN), and exacerbated
for large models which are highly over-parameterized (CIFAR10, ImageNet). This result is generally in line with observations in calibration, which point out the over-confidence of neural
network models as they become more over-parameterized \citep{guo2017calibration, wang2021rethinking}.
% \todo{make this last sentence a footnote?}

Following this, it seems that information about the penultimate layer encoding is important for high performing encoders in the ImageNet baselines. Besides VarSum, every possible $V$ improves across all ImageNet baselines when multiplied by $\norm{\textbf{h}}_1$. It is unclear why VarSum and {\egname} do not follow this trend and have subpar performance; however, we believe it may be related to how the sum of variance landscape changes with a dramatic increase in classes.

We believe that at the time of writing this paper \textsc{GradNorm} is the current state-of-the-art post hoc method for these baselines. \textit{However, importantly, we find that} ${\norm{\textbf{h}}_1 \times \textsc{Energy}}$ \textit{is a strictly better detector than \textsc{GradNorm} on the ImageNet baselines.} These results further strengthen our claim that 
gradients do not necessarily provide unique benefits
for OOD detection performance. It is perhaps only the encoding-output decomposition that results in strong performance, rather than some unique property of gradients.

% Interestingly, only using the TV distance resulted in the best AUROC on average across all settings.
% When focusing on the small-scale setting, we see that 
% only using a $V$ term tended to result in competitive performance, which is a similar pattern we observed in the previous experiments. 
% We see that the output scores decline in detection performance in the large scale setting. 
% However, by simply pairing the score with a 
% 1-norm encoding of the inputs alleviates the decline and we can achieve performance (e.g. $\norm{h}_{1} *$ Energy) that is strictly 
% better in both AUROC and FPR\textsubscript{95} than a gradient-based score (\textsc{GradNorm}).

% Rather, a much more computationally efficient and 
% versatile method of pairing an adequate input encoding ($U$) with an output probability derived score ($V$) 
% can suffice as a performant score

\subsection{Exploring Encoding Choices}
\label{sec:explore_encoding}
While the previous section limited the feature encoding
to the $L_1$ norm, in this section, we study the 
effect of encoding design choice by varying the norm 
applied. Although \citet{huang2021importance} do an ablation which changes the order of the norm, they do this ablation on the gradients themselves. By focusing on the norm of $\textbf{h}$ only, we are able to have more control over the score and the resulting detector. 

Figure~\ref{fig:norm_variation} displays the average AUROC performance 
in the ImageNet benchmark for 48 different pairs of $UV$ scores, where $U$ $\in \{\norm{\bh}_{p} : p \in \{0, 0.1, 0.3, 0.5, 0.8, 1, 2, 3, 4, 5, 6, \infty \}\}$, and $V$ $\in$ \{Energy, TV, MSP, VarSum\}. Heatmaps for every experiment are shown in Appendix~\ref{app:more_heat}.

The results indicate that the choice of encoding
does have a significant effect in a score's OOD detection performance.
In particular, $\norm{\textbf{h}}_{0.3} \times \textrm{Energy}$ achieves an AUROC of 0.917, which is a drastic improvement over any score in Table~\ref{tab:all_aucs}. It is also better than some previously proposed non-post hoc methods such as MOS \citep{huang2021mos}, which achieves an average AUROC of 0.901.

% , hence we do not 
% claim this is a new SotA baseline 
% we achieve with our own method (unlike previous works which choose their method via ablations on the test set\citep{huang2021importance}). 
We note that this result is not exactly fair since the parameter scan was done over the benchmark test set. Nevertheless, this is an encouraging result for what could possibly be achieved by a method that considers both image encoding and network output. In particular, we believe that devising more sophisticated $U$ terms than the naive ones tested here could be a promising direction for improving 
OOD detection performance within this framework.

\subsection{Challenging the Feature-Extraction Hypothesis}
As described above, one emerging theory that attempts to explain the role of gradients in OOD detection is the feature-extraction hypothesis. Quoting \citet{lee2020gradients}, 

\begin{displayquote}
\vspace{-1.7mm}
\textit{Gradient-based optimization involves larger updates when there is a
larger gap between predictions and correct labels for given inputs. It implies that the model requires more significant adjustments to its parameters, as it has not learned enough features to represent the inputs or relationships between learned features and classes for correct prediction}. 
\vspace{-1.7mm}
\end{displayquote}

We challenge the plausibility of the feature-extraction hypothesis in explaining the role of gradients by the following observation: by taking gradients at a singleton input point $\tilde{\mathbf{x}}$, {\gn} can be understood as performing a local optimization to a classification problem with a dataset of the form $\mathcal{D}= \left\{(\tilde{\mathbf{x}},Y_i)\right\}_{i=1}^{N}$ where $Y_i \sim \text{Uniform}$. Importantly, we note that fitting such a dataset does not require learning a true mapping from image space to the space of distributions over classes. To fit such a dataset well, all that is required is to find a parameterization of $\mathbf{p}(\tilde{\mathbf{x}})$ that closely aligns with the Uniform distribution at $\tilde{\mathbf{x}}$, with no constraints on outputs at other input points. We therefore hypothesize that the high performance of gradient-based OOD methods has been misattributed to their ability to detect needed changes to feature representation. 

To test this hypothesis, we re-ran 
the same small-scale benchmark experiment from Section~\ref{sec:different_grads}
with a new, carefully designed 
gradient-based score which is 
intended to align with an objective that \textit{necessarily} requires learning a true mapping from image space to distribution space to perform well. 
% Specifically, we define the score
% To test this hypothesis, we ran a similar experiment to that in Section 4.1 with a gradient-based score function designed to align with an objective that \textit{necessarily} requires learning a true mapping from image space to distribution space to perform well. 
Specifically, we define the score
\begin{align} \label{eq:batch_grad}
\nonumber
    &\mathcal{S}_{BG}({\mathbf{x}}) =
    \\
    &\mathbb{E}_{Y\sim\mathbf{p}}\left[\left\lVert \nabla_\theta  \left(\log \mathbf{p}^{(Y)}({\mathbf{x}}) + \sum_{i=1}^C \log \mathbf{p}^{(Y_i^{\text{ID}})}(\mathbf{x}_i^{\text{ID}})\right)\right\rVert_{1}\right],
\end{align}
where $(\mathbf{x}_i^{\text{ID}},Y_i^{\text{ID}})$ is an in distribution training point belonging to class $i$. We refer to the OOD classifier based on this score function as \batchgrad. The loss in Eq. \ref{eq:batch_grad} aims to answer the question: how hard is it (in terms of gradient norms) to learn a classifier that actually \textit{does} require a true mapping to perform well? If the true role of gradients in OOD detection is in feature extraction, then we would expect the score described in Eq. \ref{eq:batch_grad} to perform better for Deep variants compared to Shallow. However, as shown in the results in Table \ref{tab:batch_grad_table} we find that the \textit{opposite} holds:
the gradients of just the last layer were more informative for OOD detection than the gradients of the whole network.  
This observation is consistent with previous experiments showing the sufficiency, and at times improvement, of restricting gradients to final layer parameters.

This experimental result, in combination with analysis demonstrating the significance of the last layer gradients, leads us to conclude that the feature-extraction hypothesis is not an appropriate explanation for the high performance of gradient-based OOD detection methods.

\section{Discussion}\label{sec:conclusion}

In this work, we experimentally investigated gradient-based OOD detection methods. While using gradient-based approaches can result in strong performance, we find previous explanations attributing performance to gradients unsatisfactory.

Although prior works have focused on the interpretation of taking the gradient with respect to the KL divergence between the outputted distribution and a discrete uniform distribution \citep{lee2020gradients, huang2021importance}, we find that other gradient-based scores that do not have this interpretation \textit{also perform well}, especially on smaller scale problems.

We also question the idea that the gradient space of the neural network holds key information used for OOD detection. Our experiments provide evidence against the hypothesis that gradient-based methods are informed by large changes needed in the network to capture unseen, OOD images. Moreover, we show that we can derive better performing detectors that are agnostic to gradients and only use the encoding-output decomposition discussed in \citet{huang2021importance}. As such, we believe the strength of \textsc{GradNorm} comes not from its leverage of gradients, but solely from the fact that it fuses information about network encodings and outputted distributions. \textit{Hence, while it is possible that gradients contain useful information for OOD detection, we do not believe that previous methods leverage information from gradients that cannot be derived more easily through other means.}
\vspace{-1.7mm}
\begin{table}[t]
\centering
\begin{tabular}{|l|c|}
\cline{1-2}
\multicolumn{1}{|c|}{ } & \textbf{AUROC} \\ \hline
\textbf{Deep \batchgrad}    & 0.787 $(\pm 0.224) $                              \\ \hline
\textbf{Shallow \batchgrad} & \textbf{0.925} $(\pm 0.044)$              \\ \hline
\end{tabular}
    \caption{\textbf{\batchgrad $\,$ Small-Scale Experimental Results} This table shows AUROC results, calculated across the 6 ID-OOD dataset combinations from \{MNIST, CIFAR-10, SVHN\}.}
    \label{tab:batch_grad_table}
\end{table}

\textbf{Future Work} 
Our hope is that the insights provided in this work can be used to further improve OOD detection. 
In particular, we believe that more advanced methods can be developed that fuse together information from the network encoding and scores derived from
the model's predicted distribution. 
For example, we evaluated the choice of input encodings
by varying the order of the norm applied 
on the last hidden layer outputs.
Investigating the utility of other hidden layer features or auxiliary encoders and studying
what properties of an encoding are helpful in detecting OOD data
could provide further insights to devising stronger OOD detectors.

% utilizing an auxiliary encoder 

% Our hope is that the insights in this work can be used to further better OOD detection. In particular, we believe that more advanced methods can be developed that fuse together network encoding information with outputted probabilities. 

Another interesting direction for investigation is what role task difficulty and model capacity play in how these detectors perform. We found that when the in distribution task was easier, top performing detectors only depended on network outputs; however, it was essential to use both outputs and input encodings for the more complex ImageNet tasks. 
As mentioned in Section~\ref{sec:explore_encoding_output}, 
existing works in uncertainty quantification and calibration  
have noted the unreliability of a deep model's predicted class distribution.
While calibration is orthogonal to the scope of this work, 
we believe there could be fundamental ties in determining
the boundary between solely relying on the output predicted distribution for OOD detection (i.e. utilizing the $V$ term only), and additionally requiring extracted information from 
the input space via input encodings (utilizing the $U$ term).
We leave investigating this direction for future work.

% Insights to this
% question could help practitioners understand when to use
% different forms of OOD detection, and which aspects can
% lead to stronger detection performance.

% \begin{table*}[t]
%     \centering
%     \begin{tabular}{| c c | c c | c c | c c | c c | c c |} \hline
%      && \multicolumn{2}{|c|}{\textbf{iNaturalist}} & \multicolumn{2}{|c|}{\textbf{SUN}} & \multicolumn{2}{|c|}{\textbf{Places}} & \multicolumn{2}{|c|}{\textbf{Textures}} & \multicolumn{2}{|c|}{\textbf{Average}} \\
%     \hline
%     \multicolumn{2}{|c|}{\textbf{Method}} & FPR95 & AUROC & FPR95 & AUROC & FPR95 & AUROC & FPR95 & AUROC & FPR95 & AUROC \\
%     && $\downarrow$ & $\uparrow$ & $\downarrow$ & $\uparrow$ & $\downarrow$ & $\uparrow$ & $\downarrow$ & $\uparrow$ & $\downarrow$ & $\uparrow$ \\
%     \hline
%     \multicolumn{2}{|l|}{MSP (REFERENCE)} & 63.69 & 87.59 & 79.98 & 78.34 & 81.44 & 76.76 & 82.73 & 74.45 & 76.96 & 79.29 \\
%     \multicolumn{2}{|l|}{ODIN (REFERENCE)} & 62.69 & 89.36 & 71.67 & 83.92 & 76.27 & 80.67 & 81.31 & 76.30 & 72.99 & 82.56 \\
%     \multicolumn{2}{|l|}{Energy (REFERENCE)} & 64.91 & 88.48 & 65.33 & 85.32 & 73.02 & 81.37 & 80.87 & 75.79 & 71.03 & 82.74 \\
%     \multicolumn{2}{|l|}{Mahalanobis (REFERENCE)} & 96.35 & 46.33 & 88.43 & 65.20 & 89.76 & 64.46 & 52.23 & 72.10 & 81.69 & 62.02 \\
%     \multicolumn{2}{|l|}{Mahalanobis (REFERENCE)} & 96.35 & 46.33 & 88.43 & 65.20 & 89.76 & 64.46 & 52.23 & 72.10 & 81.69 & 62.02 \\
%     \end{tabular}
%     \caption{Caption}
%     \label{tab:my_label}
% \end{table*}

\section{Acknowledgments}

This material is based upon work supported by the National Science Foundation Graduate Research
Fellowship Program under Grant No. DGE1745016/DGE2140739. Any opinions, findings, and conclusions
or recommendations expressed in this material are those of the author(s) and do not necessarily reflect
the views of the National Science Foundation.

Youngseog Chung is supported by the Kwanjeong Educational Foundation.

\newpage
\nocite{langley00}

\bibliography{refs}

\begin{thebibliography}{40}
\providecommand{\natexlab}[1]{#1}
\providecommand{\url}[1]{\texttt{#1}}
\expandafter\ifx\csname urlstyle\endcsname\relax
  \providecommand{\doi}[1]{doi: #1}\else
  \providecommand{\doi}{doi: \begingroup \urlstyle{rm}\Url}\fi

\bibitem[Agarwal et~al.(2020)Agarwal, D'souza, and
  Hooker]{agarwal2020estimating}
Agarwal, C., D'souza, D., and Hooker, S.
\newblock Estimating example difficulty using variance of gradients.
\newblock \emph{arXiv preprint arXiv:2008.11600}, 2020.

\bibitem[Agarwal et~al.(2021)Agarwal, Arora, and
  Schneider]{agarwal2021learning}
Agarwal, T., Arora, H., and Schneider, J.
\newblock Learning urban driving policies using deep reinforcement learning.
\newblock In \emph{2021 IEEE International Intelligent Transportation Systems
  Conference (ITSC)}, pp.\  607--614. IEEE, 2021.

\bibitem[Boyer et~al.(2021)Boyer, Wai, Clement, Kolemen, Char, Chung,
  Neiswanger, and Schneider]{boyer2021machine}
Boyer, M., Wai, J., Clement, M., Kolemen, E., Char, I., Chung, Y., Neiswanger,
  W., and Schneider, J.
\newblock Machine learning for tokamak scenario optimization: combining
  accelerating physics models and empirical models.
\newblock \emph{Bulletin of the American Physical Society}, 2021.

\bibitem[Char et~al.(2021)Char, Chung, Boyer, Kolemen, and
  Schneider]{char2021model}
Char, I., Chung, Y., Boyer, M., Kolemen, E., and Schneider, J.
\newblock A model-based reinforcement learning approach for beta control.
\newblock In \emph{APS Division of Plasma Physics Meeting Abstracts}, volume
  2021, pp.\  PP11--150, 2021.

\bibitem[Cimpoi et~al.(2014)Cimpoi, Maji, Kokkinos, Mohamed, and
  Vedaldi]{cimpoi2014describing}
Cimpoi, M., Maji, S., Kokkinos, I., Mohamed, S., and Vedaldi, A.
\newblock Describing textures in the wild.
\newblock In \emph{Proceedings of the IEEE Conference on Computer Vision and
  Pattern Recognition}, pp.\  3606--3613, 2014.

\bibitem[Deng et~al.(2009)Deng, Dong, Socher, Li, Li, and
  Fei-Fei]{deng2009imagenet}
Deng, J., Dong, W., Socher, R., Li, L.-J., Li, K., and Fei-Fei, L.
\newblock Imagenet: A large-scale hierarchical image database.
\newblock In \emph{2009 IEEE conference on computer vision and pattern
  recognition}, pp.\  248--255. Ieee, 2009.

\bibitem[Deng(2012)]{deng2012mnist}
Deng, L.
\newblock The mnist database of handwritten digit images for machine learning
  research.
\newblock \emph{IEEE Signal Processing Magazine}, 29\penalty0 (6):\penalty0
  141--142, 2012.

\bibitem[Guo et~al.(2017)Guo, Pleiss, Sun, and Weinberger]{guo2017calibration}
Guo, C., Pleiss, G., Sun, Y., and Weinberger, K.~Q.
\newblock On calibration of modern neural networks.
\newblock In \emph{International Conference on Machine Learning}, pp.\
  1321--1330. PMLR, 2017.

\bibitem[He et~al.(2016)He, Zhang, Ren, and Sun]{he2016identity}
He, K., Zhang, X., Ren, S., and Sun, J.
\newblock Identity mappings in deep residual networks.
\newblock In \emph{European conference on computer vision}, pp.\  630--645.
  Springer, 2016.

\bibitem[Hein et~al.(2019)Hein, Andriushchenko, and Bitterwolf]{Hein_2019_CVPR}
Hein, M., Andriushchenko, M., and Bitterwolf, J.
\newblock Why relu networks yield high-confidence predictions far away from the
  training data and how to mitigate the problem.
\newblock In \emph{Proceedings of the IEEE/CVF Conference on Computer Vision
  and Pattern Recognition (CVPR)}, June 2019.

\bibitem[Hendrycks \& Gimpel(2016)Hendrycks and Gimpel]{hendrycks2016baseline}
Hendrycks, D. and Gimpel, K.
\newblock A baseline for detecting misclassified and out-of-distribution
  examples in neural networks.
\newblock \emph{arXiv preprint arXiv:1610.02136}, 2016.

\bibitem[Hendrycks et~al.(2018)Hendrycks, Mazeika, and
  Dietterich]{hendrycks2018deep}
Hendrycks, D., Mazeika, M., and Dietterich, T.
\newblock Deep anomaly detection with outlier exposure.
\newblock \emph{arXiv preprint arXiv:1812.04606}, 2018.

\bibitem[Huang \& Li(2021)Huang and Li]{huang2021mos}
Huang, R. and Li, Y.
\newblock Mos: Towards scaling out-of-distribution detection for large semantic
  space.
\newblock In \emph{Proceedings of the IEEE/CVF Conference on Computer Vision
  and Pattern Recognition}, pp.\  8710--8719, 2021.

\bibitem[Huang et~al.(2021)Huang, Geng, and Li]{huang2021importance}
Huang, R., Geng, A., and Li, Y.
\newblock On the importance of gradients for detecting distributional shifts in
  the wild.
\newblock \emph{Advances in Neural Information Processing Systems}, 34, 2021.

\bibitem[Kirsch et~al.(2021)Kirsch, Mukhoti, Amersfoort, Torr, and
  Gal]{kirsch2021pitfall}
Kirsch, A., Mukhoti, J., Amersfoort, J.~v., Torr, P.~H., and Gal, Y.
\newblock On pitfalls in ood detection: Predictive entropy considered harmful.
\newblock 2021.

\bibitem[Kolesnikov et~al.(2020)Kolesnikov, Beyer, Zhai, Puigcerver, Yung,
  Gelly, and Houlsby]{kolesnikov2020big}
Kolesnikov, A., Beyer, L., Zhai, X., Puigcerver, J., Yung, J., Gelly, S., and
  Houlsby, N.
\newblock Big transfer (bit): General visual representation learning.
\newblock In \emph{Computer Vision--ECCV 2020: 16th European Conference,
  Glasgow, UK, August 23--28, 2020, Proceedings, Part V 16}, pp.\  491--507.
  Springer, 2020.

\bibitem[Krizhevsky(2009)]{Krizhevsky09learningmultiple}
Krizhevsky, A.
\newblock Learning multiple layers of features from tiny images.
\newblock Technical report, 2009.

\bibitem[Lakshminarayanan et~al.(2016)Lakshminarayanan, Pritzel, and
  Blundell]{lakshminarayanan2016simple}
Lakshminarayanan, B., Pritzel, A., and Blundell, C.
\newblock Simple and scalable predictive uncertainty estimation using deep
  ensembles.
\newblock \emph{arXiv preprint arXiv:1612.01474}, 2016.

\bibitem[Lee \& AlRegib(2020)Lee and AlRegib]{lee2020gradients}
Lee, J. and AlRegib, G.
\newblock Gradients as a measure of uncertainty in neural networks.
\newblock In \emph{2020 IEEE International Conference on Image Processing
  (ICIP)}, pp.\  2416--2420. IEEE, 2020.

\bibitem[Lee et~al.(2017)Lee, Lee, Lee, and Shin]{lee2017training}
Lee, K., Lee, H., Lee, K., and Shin, J.
\newblock Training confidence-calibrated classifiers for detecting
  out-of-distribution samples.
\newblock \emph{arXiv preprint arXiv:1711.09325}, 2017.

\bibitem[Lee et~al.(2018)Lee, Lee, Lee, and Shin]{lee2018simple}
Lee, K., Lee, K., Lee, H., and Shin, J.
\newblock A simple unified framework for detecting out-of-distribution samples
  and adversarial attacks.
\newblock \emph{Advances in neural information processing systems}, 31, 2018.

\bibitem[Liang et~al.(2017)Liang, Li, and Srikant]{liang2017enhancing}
Liang, S., Li, Y., and Srikant, R.
\newblock Enhancing the reliability of out-of-distribution image detection in
  neural networks.
\newblock \emph{arXiv preprint arXiv:1706.02690}, 2017.

\bibitem[Linmans et~al.(2020)Linmans, van~der Laak, and
  Litjens]{linmans2020efficient}
Linmans, J., van~der Laak, J., and Litjens, G.
\newblock Efficient out-of-distribution detection in digital pathology using
  multi-head convolutional neural networks.
\newblock In \emph{MIDL}, pp.\  465--478, 2020.

\bibitem[Liu et~al.(2020)Liu, Wang, Owens, and Li]{liu2020energy}
Liu, W., Wang, X., Owens, J.~D., and Li, Y.
\newblock Energy-based out-of-distribution detection.
\newblock \emph{arXiv preprint arXiv:2010.03759}, 2020.

\bibitem[Netzer et~al.(2011)Netzer, Wang, Coates, Bissacco, Wu, and
  Ng]{Netzer2011ReadingDI}
Netzer, Y., Wang, T., Coates, A., Bissacco, A., Wu, B., and Ng, A.
\newblock Reading digits in natural images with unsupervised feature learning.
\newblock 2011.

\bibitem[Ren et~al.(2019)Ren, Liu, Fertig, Snoek, Poplin, DePristo, Dillon, and
  Lakshminarayanan]{ren2019likelihood}
Ren, J., Liu, P.~J., Fertig, E., Snoek, J., Poplin, R., DePristo, M.~A.,
  Dillon, J.~V., and Lakshminarayanan, B.
\newblock Likelihood ratios for out-of-distribution detection.
\newblock \emph{arXiv preprint arXiv:1906.02845}, 2019.

\bibitem[Rudin(2019)]{rudin2019stop}
Rudin, C.
\newblock Stop explaining black box machine learning models for high stakes
  decisions and use interpretable models instead.
\newblock \emph{Nature Machine Intelligence}, 1\penalty0 (5):\penalty0
  206--215, 2019.

\bibitem[Salehi et~al.(2021)Salehi, Mirzaei, Hendrycks, Li, Rohban, and
  Sabokrou]{salehi2021unified}
Salehi, M., Mirzaei, H., Hendrycks, D., Li, Y., Rohban, M.~H., and Sabokrou, M.
\newblock A unified survey on anomaly, novelty, open-set, and
  out-of-distribution detection: Solutions and future challenges.
\newblock \emph{arXiv preprint arXiv:2110.14051}, 2021.

\bibitem[Serr{\`a} et~al.(2019)Serr{\`a}, {\'A}lvarez, G{\'o}mez, Slizovskaia,
  N{\'u}{\~n}ez, and Luque]{serra2019input}
Serr{\`a}, J., {\'A}lvarez, D., G{\'o}mez, V., Slizovskaia, O., N{\'u}{\~n}ez,
  J.~F., and Luque, J.
\newblock Input complexity and out-of-distribution detection with
  likelihood-based generative models.
\newblock \emph{arXiv preprint arXiv:1909.11480}, 2019.

\bibitem[Techapanurak et~al.(2020)Techapanurak, Suganuma, and
  Okatani]{techapanurak2020hyperparameter}
Techapanurak, E., Suganuma, M., and Okatani, T.
\newblock Hyperparameter-free out-of-distribution detection using cosine
  similarity.
\newblock In \emph{Proceedings of the Asian Conference on Computer Vision},
  2020.

\bibitem[Van~Amersfoort et~al.(2020)Van~Amersfoort, Smith, Teh, and
  Gal]{van2020uncertainty}
Van~Amersfoort, J., Smith, L., Teh, Y.~W., and Gal, Y.
\newblock Uncertainty estimation using a single deep deterministic neural
  network.
\newblock In \emph{International Conference on Machine Learning}, pp.\
  9690--9700. PMLR, 2020.

\bibitem[Van~Horn et~al.(2017)Van~Horn, Mac~Aodha, Song, Cui, Sun, Shepard,
  Adam, Perona, and Belongie]{van2017inaturalist}
Van~Horn, G., Mac~Aodha, O., Song, Y., Cui, Y., Sun, C., Shepard, A., Adam, H.,
  Perona, P., and Belongie, S.
\newblock The inaturalist species classification and detection
  dataset-supplementary material.
\newblock \emph{Reptilia}, 32\penalty0 (400):\penalty0 1--3, 2017.

\bibitem[Vernekar et~al.(2019)Vernekar, Gaurav, Abdelzad, Denouden, Salay, and
  Czarnecki]{vernekar2019out}
Vernekar, S., Gaurav, A., Abdelzad, V., Denouden, T., Salay, R., and Czarnecki,
  K.
\newblock Out-of-distribution detection in classifiers via generation.
\newblock \emph{arXiv preprint arXiv:1910.04241}, 2019.

\bibitem[Wang et~al.(2021)Wang, Feng, and Zhang]{wang2021rethinking}
Wang, D.-B., Feng, L., and Zhang, M.-L.
\newblock Rethinking calibration of deep neural networks: Do not be afraid of
  overconfidence.
\newblock \emph{Advances in Neural Information Processing Systems}, 34, 2021.

\bibitem[Xiao et~al.(2010)Xiao, Hays, Ehinger, Oliva, and
  Torralba]{xiao2010sun}
Xiao, J., Hays, J., Ehinger, K.~A., Oliva, A., and Torralba, A.
\newblock Sun database: Large-scale scene recognition from abbey to zoo.
\newblock In \emph{2010 IEEE computer society conference on computer vision and
  pattern recognition}, pp.\  3485--3492. IEEE, 2010.

\bibitem[Yang et~al.(2021)Yang, Zhou, Li, and Liu]{yang2021generalized}
Yang, J., Zhou, K., Li, Y., and Liu, Z.
\newblock Generalized out-of-distribution detection: A survey.
\newblock \emph{arXiv preprint arXiv:2110.11334}, 2021.

\bibitem[Yu \& Aizawa(2019)Yu and Aizawa]{Yu_2019_ICCV}
Yu, Q. and Aizawa, K.
\newblock Unsupervised out-of-distribution detection by maximum classifier
  discrepancy.
\newblock In \emph{Proceedings of the IEEE/CVF International Conference on
  Computer Vision (ICCV)}, October 2019.

\bibitem[Zhou et~al.(2017)Zhou, Lapedriza, Khosla, Oliva, and
  Torralba]{zhou2017places}
Zhou, B., Lapedriza, A., Khosla, A., Oliva, A., and Torralba, A.
\newblock Places: A 10 million image database for scene recognition.
\newblock \emph{IEEE transactions on pattern analysis and machine
  intelligence}, 40\penalty0 (6):\penalty0 1452--1464, 2017.

\bibitem[Zhou et~al.(2020)Zhou, Cheng, Lipton, Chen, and
  Weiss]{zhou2020mortality}
Zhou, H., Cheng, C., Lipton, Z.~C., Chen, G.~H., and Weiss, J.~C.
\newblock Mortality risk score for critically ill patients with viral or
  unspecified pneumonia: Assisting clinicians with covid-19 ecmo planning.
\newblock In \emph{International Conference on Artificial Intelligence in
  Medicine}, pp.\  336--347. Springer, 2020.

\bibitem[Zisselman \& Tamar(2020)Zisselman and Tamar]{zisselman2020deep}
Zisselman, E. and Tamar, A.
\newblock Deep residual flow for out of distribution detection.
\newblock In \emph{Proceedings of the IEEE/CVF Conference on Computer Vision
  and Pattern Recognition}, pp.\  13994--14003, 2020.

\end{thebibliography}
\bibliographystyle{icml2022}

\newpage
%%%%%%%%%%%%%%%%%%%%%%%%%%%%%%%%%%%%%%%%%%%%%%%%%%%%%%%%%%%%%%%%%%%%%%%%%%%%%%%
%%%%%%%%%%%%%%%%%%%%%%%%%%%%%%%%%%%%%%%%%%%%%%%%%%%%%%%%%%%%%%%%%%%%%%%%%%%%%%%
% APPENDIX
%%%%%%%%%%%%%%%%%%%%%%%%%%%%%%%%%%%%%%%%%%%%%%%%%%%%%%%%%%%%%%%%%%%%%%%%%%%%%%%
%%%%%%%%%%%%%%%%%%%%%%%%%%%%%%%%%%%%%%%%%%%%%%%%%%%%%%%%%%%%%%%%%%%%%%%%%%%%%%%
\newpage
\appendix
\onecolumn
\section{Additional Experimental Details}
\label{app:tables}


% \usepackage{multirow}
The tables below shows detailed breakdowns of AUROC results for the small-scale experiments. Note that the top row describes the ID dataset, and the two entries beneath describe the OOD datasets.
\begin{table}[h]
\renewcommand{\arraystretch}{1.5}
\centering
\begin{tabular}{|llllllll|} 
\hline
\multirow{2}{*}{\textbf{Score Expression }}                                                                                                      & \multirow{2}{*}{\textbf{Gradient }} & \multicolumn{2}{c}{\textbf{SVHN }} & \multicolumn{2}{c}{\textbf{MNIST }} & \multicolumn{2}{c|}{\textbf{CIFAR }}  \\
                                                                                                                                                 &                                           & \textbf{MNIST} & \textbf{CIFAR}    & \textbf{SVHN} & \textbf{CIFAR}      & \textbf{SVHN} & \textbf{MNIST}        \\ 
\hline
$\lVert \mathbb{E}_{Y\sim \mathbf{p}}\left[ \nabla_\theta\log\mathbf{p}^{(Y)} \right] \rVert_2^2$                                                & Deep                                      & 0.595          & 0.730             & 0.839         & 0.786               & 0.669         & 0.733                 \\
$\lVert \mathbb{E}_{Y\sim \mathbf{p}}\left[ \nabla_\theta\log\mathbf{p}^{(Y)} \right] \rVert_2^2$                                                & Shallow                                   & 0.651          & 0.796             & 0.838         & 0.775               & 0.635         & 0.749                 \\
$\mathbb{E}_{Y\sim \text{Uniform}}\left[\lVert\nabla_\theta\log { \mathbf{p}^{(Y)}} \rVert_1 \right]$                                            & Deep                                      & 0.692          & 0.816             & 0.790         & 0.717               & 0.993         & 0.940                 \\
$\mathbb{E}_{Y\sim \text{Uniform}}\left[\lVert\nabla_\theta\log { \mathbf{p}^{(Y)}} \rVert_1 \right]$                                            & Shallow                                   & 0.606          & 0.817             & 0.939         & 0.857               & 0.995         & 0.886                 \\
$\mathbb{E}_{Y\sim \text{Uniform}}\left[\lVert\nabla_\theta\log { \mathbf{p}^{(Y)}} \rVert_2^2 \right]$                                          & Deep                                      & 0.591          & 0.895             & 0.986         & 0.957               & 0.954         & 0.819                 \\
$\mathbb{E}_{Y\sim \text{Uniform}}\left[\lVert\nabla_\theta\log { \mathbf{p}^{(Y)}} \rVert_2^2 \right]$                                          & Shallow                                   & 0.709          & 0.919             & 0.976         & 0.903               & 0.994         & 0.819                 \\
$\left\lVert \nabla_\theta \mathbb{E}_{Y\sim \text{Uniform}}\left[\log{\mathbf{p}^{(Y)}}\right]\right\rVert_1$                                   & Deep                                      & 0.759          & 0.937             & 0.911         & 0.814               & 0.989         & 0.943                 \\
$\left\lVert \nabla_\theta \mathbb{E}_{Y\sim \text{Uniform}}\left[\log{\mathbf{p}^{(Y)}}\right]\right\rVert_1$                                   & Shallow                                   & 0.735          & 0.926             & 0.971         & 0.896               & 0.995         & 0.913                 \\
$\mathbb{E}_{Y\sim \mathbf{p}} \left[ \frac{\log \mathbf{p}^{(Y)}}{\mathbf{p}^{(Y)}} \lVert\nabla_\theta\log\mathbf{p}^{(Y)} \rVert_2^2 \right]$ & Shallow                                   & 0.749          & 0.933             & 0.983         & 0.928               & 0.994         & 0.870                 \\
$\mathbb{E}_{Y\sim \mathbf{p}}\left[\lVert\nabla_\theta\log { \mathbf{p}^{(Y)}} \rVert_2^2 \right]$                                               & Shallow                                   & 0.848          & 0.908             & 0.964         & 0.955               & 0.915         & 0.925                 \\
$\mathbb{E}_{Y\sim \mathbf{p}}\left[\lVert\nabla_\theta\log { \mathbf{p}^{(Y)}} \rVert_2^2 \right]$                                              & Deep                                      & 0.862          & 0.936             & 0.956         & 0.947               & 0.916         & 0.910                 \\
$\mathbb{E}_{Y\sim \mathbf{p}} \left[ \frac{\log \mathbf{p}^{(Y)}}{\mathbf{p}^{(Y)}} \lVert\nabla_\theta\log\mathbf{p}^{(Y)} \rVert_2^2 \right]$ & Deep                                      & 0.710          & 0.926             & 0.993         & 0.968               & 0.984         & 0.947                 \\
$\mathbb{E}_{Y\sim \mathbf{p}}\left[\lVert\nabla_\theta\log { \mathbf{p}^{(Y)}} \rVert_1 \right]$                                                & Shallow                                   & 0.840          & 0.931             & 0.975         & 0.963               & 0.919         & 0.921                 \\
$\mathbb{E}_{Y\sim \mathbf{p}}\left[\lVert\nabla_\theta\log { \mathbf{p}^{(Y)}} \rVert_1 \right]$                                                & Deep                                      & 0.830          & 0.947             & 0.977         & 0.965               & 0.921         & 0.918                 \\
$\sum_{k=1}^C \textbf{p}^{(k)} (1 - \textbf{p}^{(k)}) $                                                          &\textit{ N/A }                                      & 0.813          & 0.956             & 0.988         & 0.972               & 0.944         & 0.926                 \\
$\sum_{k=1}^C \left| \frac{1}{C} - \textbf{p}^{(k)} \right|$                                                           & \textit{N/A}                                       & 0.811          & 0.955             & 0.990         & 0.974               & 0.948         & 0.928                 \\
\hline
\end{tabular}
\end{table}

\begin{table}[h]
\centering
\begin{tabular}{|lllllll|} 
\hline
\textbf{ID Dataset}                     & \multicolumn{2}{c}{\textbf{SVHN}}                                        & \multicolumn{2}{c}{\textbf{MNIST}}                                       & \multicolumn{2}{c|}{\textbf{CIFAR }}                                       \\
\textbf{OOD Dataset}                    & \multicolumn{1}{c}{\textbf{MNIST}} & \multicolumn{1}{c}{\textbf{~CIFAR}} & \multicolumn{1}{c}{\textbf{~SVHN~}} & \multicolumn{1}{c}{\textbf{CIFAR}} & \multicolumn{1}{c}{\textbf{~SVHN}} & \multicolumn{1}{c|}{\textbf{~MNIST}}  \\ 
\hline
\multicolumn{1}{|l}{Deep  BATCHGRAD}    & 0.843                              & 0.947                               & 0.974                               & 0.950                              & 0.505                              & 0.502                                 \\
\multicolumn{1}{|l}{Shallow  BATCHGRAD} & 0.851                              & 0.922                               & 0.975                               & 0.963                              & 0.920                              & 0.921                                 \\
\hline
\end{tabular}
\end{table}

%%%%%%%%%%%%%%%%%%%%%%%%%%%%%%%%%%%%%%%%%%%%%%%%%%%%%%%%%%%%%%%%%%%%%%%%%%%%%%%
%%%%%%%%%%%%%%%%%%%%%%%%%%%%%%%%%%%%%%%%%%%%%%%%%%%%%%%%%%%%%%%%%%%%%%%%%%%%%%%

\section{Derivation of {\egname}  Decomposition}
\label{app:epi_derivation}

We will now rewrite Eq. \ref{eq:epigrad} into a more digestible form. Assume that we only take the derivative with respect to the last layer of the network, assume that $W \in \R^{C \times D}$ are the parameters for the last layer, and let the output of the last layer be $f_\theta(\textbf{h}) = W \textbf{h}$, where $\textbf{h}$ is the encoding taken in by the last layer. Although we have not considered bias terms in this formulation, note that they can easily be included by adding another dimension to $\textbf{h}$ with the value $1$. For readability, we write $\textbf{p}$ instead of $\textbf{p}(\textbf{h})$ and $f^{(k)}$ instead of $f^{(k)}(\textbf{h})$.

First, fix a class $k$, and note that
\begin{align*}
    \frac{\partial}{\partial f_\theta^{(k)}} \log \left(\frac{e^{f_\theta^{(k)}/T}}{\sum_{k' = 1}^C e^{f_\theta^{(k')/T}}}\right) &= \frac{\partial}{\partial f_\theta^{(k)}} \left(
    \frac{f_\theta^{(k)}}{T} - \log \sum_k e^{f_\theta^{(k')}/T}\right) \\
    &= \frac{1}{T}\left(1 - \frac{e^{f_\theta^{(k)}}}{\sum_k e^{f_\theta^{(k')}}}\right) \\
    &= \frac{1}{T} \left(1 - \textbf{p}^{(k)}\right)
\end{align*}
For a class $k' \neq k$,
\begin{align*}
    \frac{\partial}{\partial f_\theta^{(k')}} \log \left(\frac{e^{f_\theta^{(k)}}}{\sum_{k' = 1}^C e^{f_\theta^{(k')}}}\right) &= - \frac{e^{f_\theta^{(k')}}/T}{\sum_k e^{f_\theta^{(k')}/T}} \\
    &= \frac{-\textbf{p}^{(k')}}{T}
\end{align*}
Building on this,
\begin{align*}
    \frac{\partial}{\partial W} \log \left(\frac{e^{f_\theta^{(k)}/T}}{\sum_{k' = 1}^C e^{f_\theta^{(k')}/T}}\right)
    &= \frac{1}{T}\textbf{h} \begin{bmatrix}
    -\textbf{p}^{(1)}, & -\textbf{p}^{(2)}, & \hdots, & 1 - \textbf{p}^{(k)}, & \hdots
    \end{bmatrix} \\
    &= \frac{1}{T} \begin{bmatrix}
    -\textbf{p}^{(1)} \textbf{h}^{(1)} & -\textbf{p}^{(2)} \textbf{h}^{(1)} & \hdots & (1 - \textbf{p}^{(k)}) \textbf{h}^{(1)} & \hdots \\
    -\textbf{p}^{(1)} \textbf{h}^{(2)} & -\textbf{p}^{(2)} \textbf{h}^{(2)} & \hdots & (1 - \textbf{p}^{(k)}) \textbf{h}^{(2)} & \hdots \\
    \vdots & \vdots & \vdots & \vdots & \vdots
    \end{bmatrix}
\end{align*}

The $L1$ norm of the gradient with respect to $W$ is just the sum of the absolute values of each entry.
\begin{align*}
    \sum_{i = 1}^D \left((1 - \textbf{p}^{(k)}) |\textbf{h}_i| + \sum_{k' \neq k} \textbf{p}^{(k')} |\textbf{h}_i| \right) &= \sum_{i = 1}^D \left((1 - \textbf{p}^{(k)}) |\textbf{h}_i| + (1 - \textbf{p}^{(k)}) |\textbf{h}_i| \right) \\
    &= 2 (1 - \textbf{p}^{(k)}) \norm{\textbf{h}}_1
\end{align*}
Putting it all together,
\begin{align*}
    \mathcal{S}(x) &= \E_{k \sim \textbf{p}}\left[\norm{\nabla_W \log \left(\frac{e^{f_\theta^{(k)}/T}}{\sum_{k' = 1}^C e^{f_\theta^{(k')}/T}}\right)}_1 \right]\\
    &= \frac{2}{T}\E_{k \sim \textbf{p}}\left[(1 - \textbf{p}^{(k)}) \norm{\textbf{h}}_1 \right] \\
    &= \frac{2}{T} \norm{\textbf{h}}_1 \sum_{c = 1}^C \textbf{p}^{(k)} (1 - \textbf{p}^{(k)})
\end{align*}

\section{Additional Heatmaps}\label{app:more_heat}

In this section, we present additional heatmaps (similar to Figure~\ref{fig:norm_variation}) which indicate the AUROC for each of the methods evaluated in Section~\ref{sec:explore_encoding}, separately for each of the ID-OOD dataset settings tested in the experiments section.
The title of each plot indicates the ID and OOD dataset in order, i.e. ``ID x OOD''.
Each cell of the heatmap shows the AUROC for a different configuration of probability output score (y-axis) and order of the norm on the encoding fed to the last layer (x-axis).

%%%% SVHN - 
\begin{figure}[h!]
    \centering
    \includegraphics[width=1\linewidth]{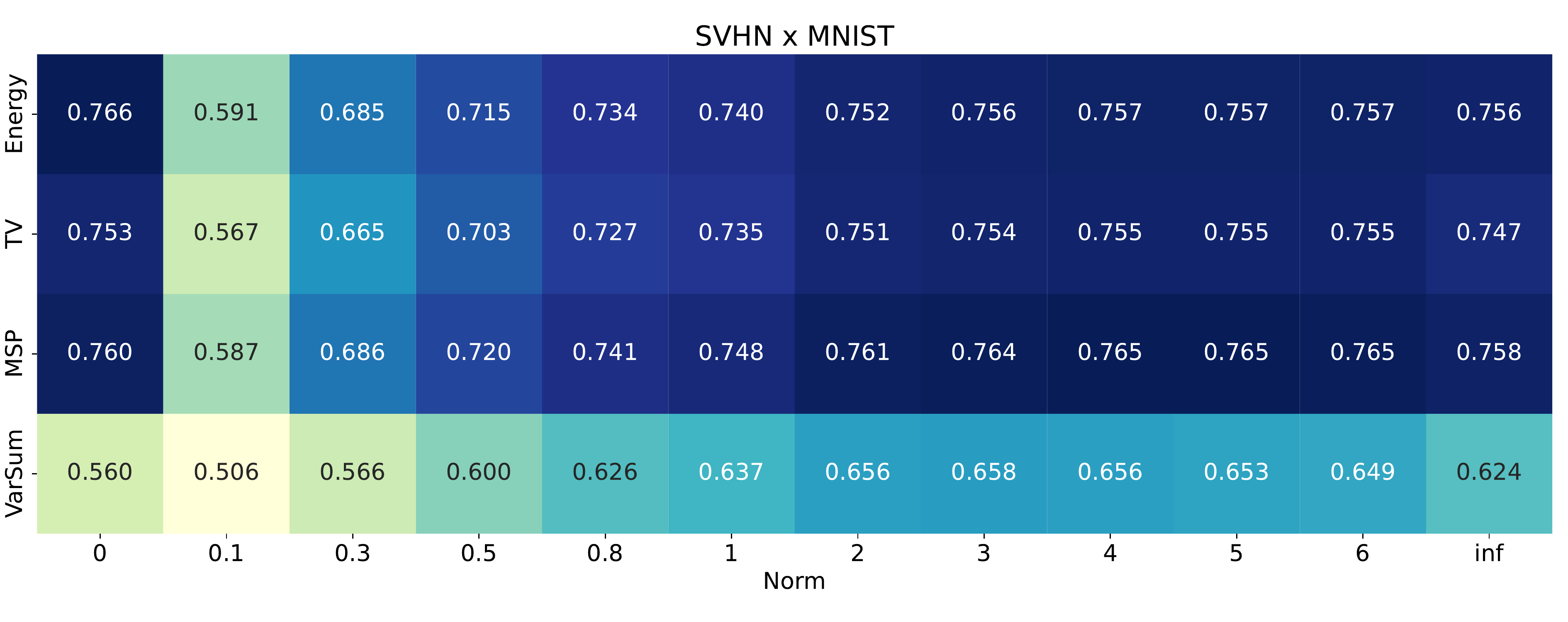}
    \vspace{-5mm}
    % \caption{\textbf{AUROC on SVHN (ID) - MNIST (OOD) Experiment.} Each cell of the heatmap shows the AUROC for a different configuration of probability output score (y-axis) and order of the norm on the encoding fed to the last layer (x-axis).}
    % \label{fig:svhn_mnist_heatmap}
\end{figure}

\begin{figure}[h!]
    \centering
    \includegraphics[width=1\linewidth]{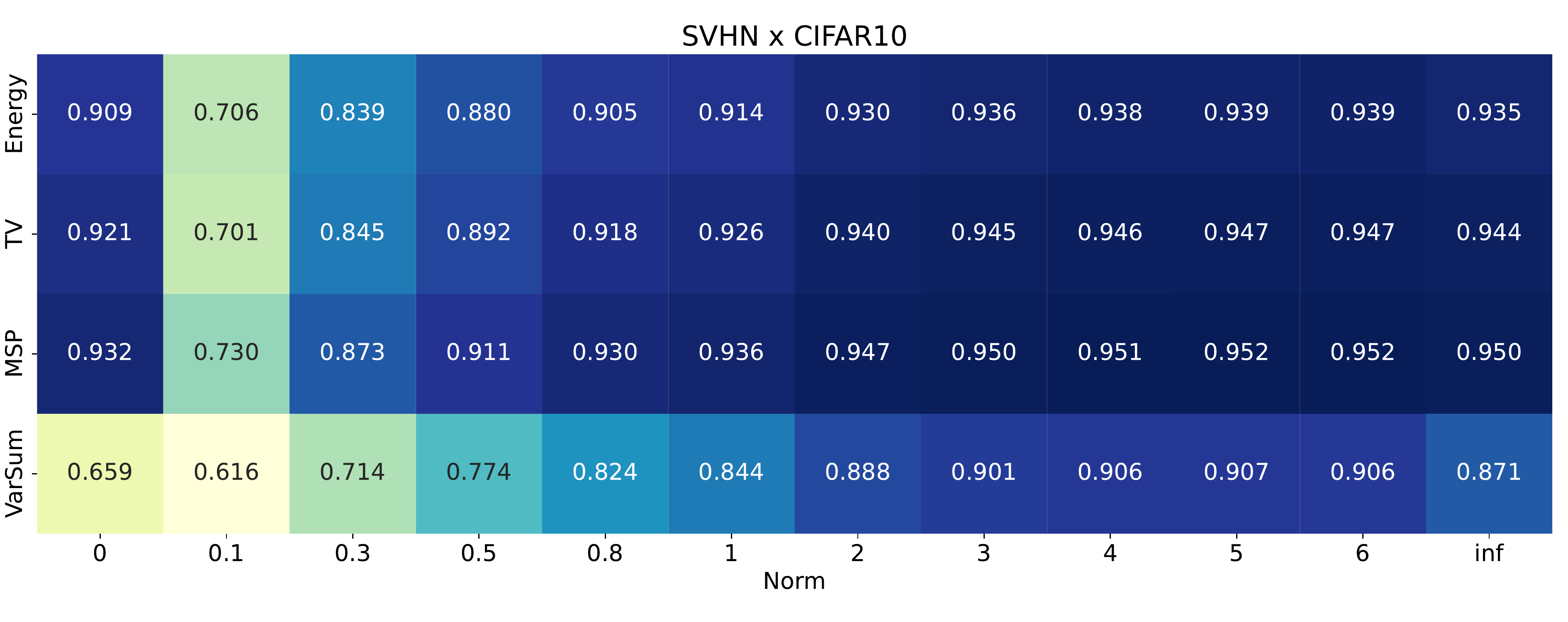}
    % \label{fig:my_label}
\end{figure}

%%%% MNIST - 
\begin{figure}[h!]
    \centering
    \includegraphics[width=1\linewidth]{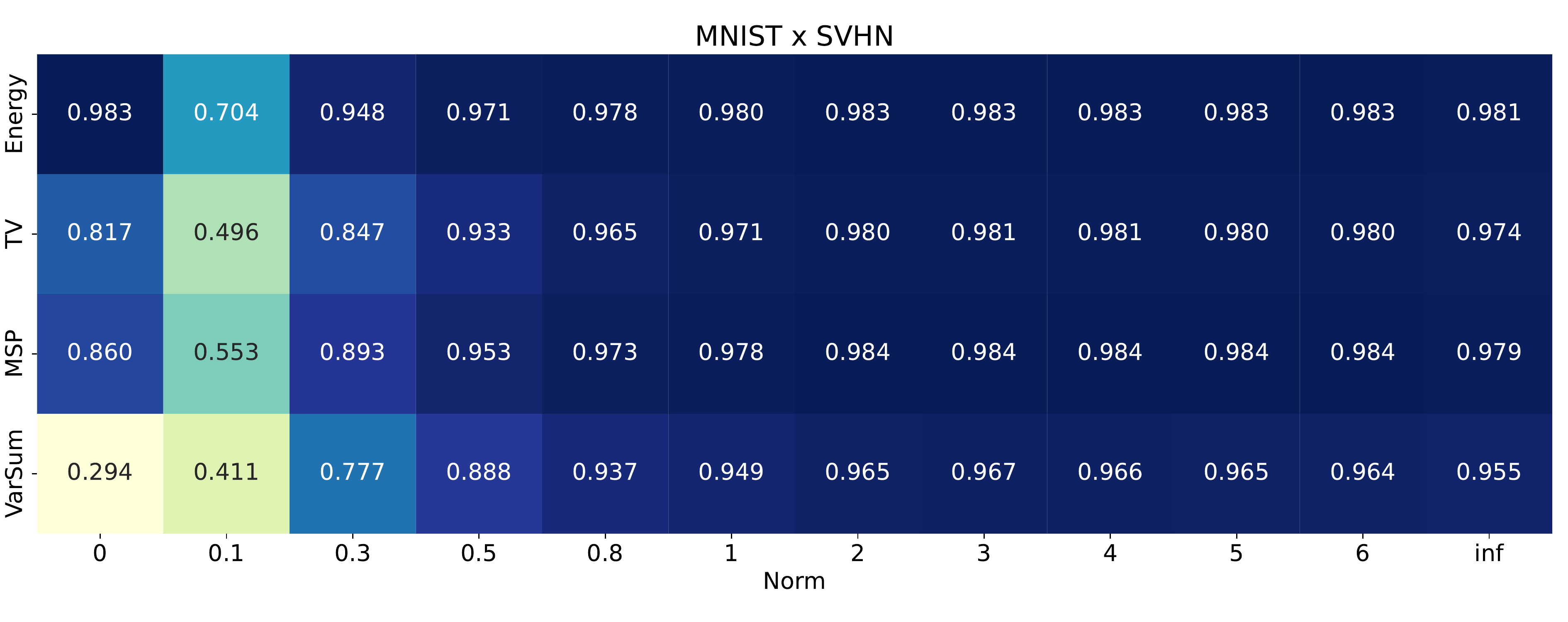}
    % \label{fig:my_label}
\end{figure}
\begin{figure}[h!]
    \centering
    \includegraphics[width=1\linewidth]{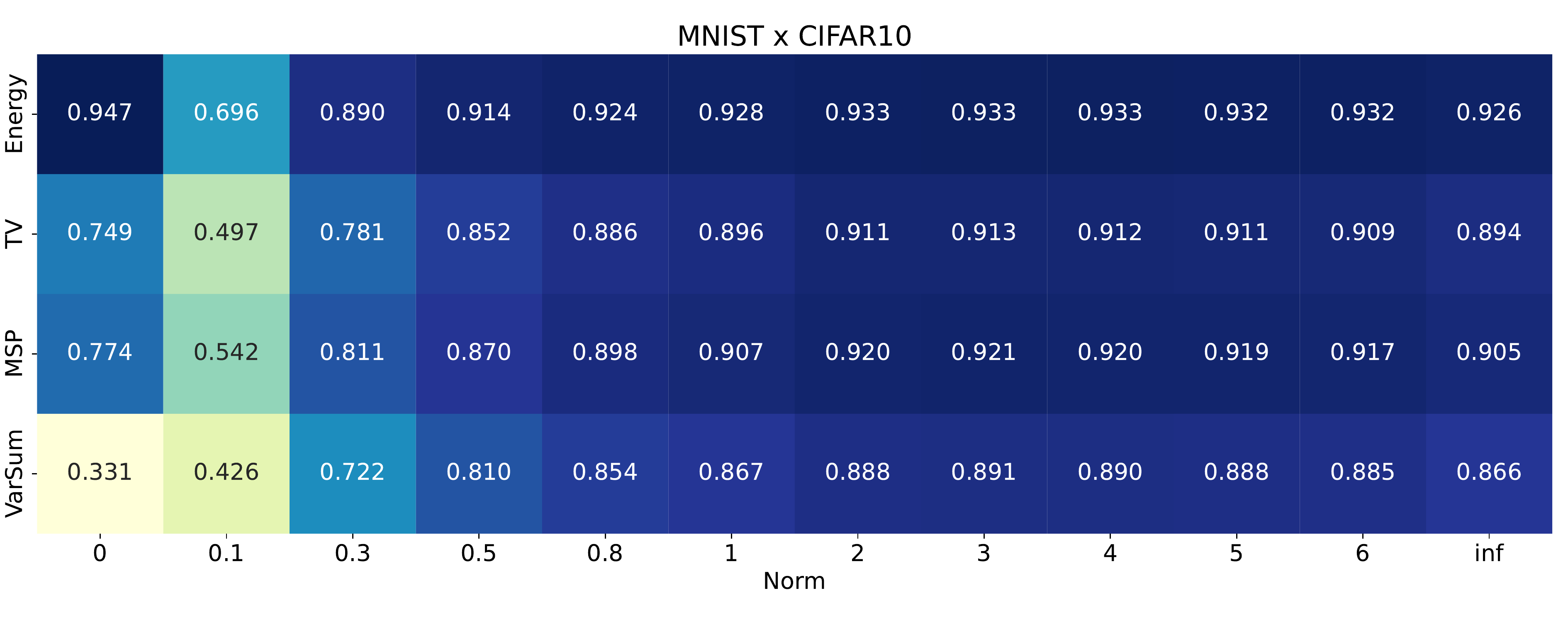}
    % \label{fig:my_label}
\end{figure}

%%%% CIFAR - 
\begin{figure}[h!]
    \centering
    \includegraphics[width=1\linewidth]{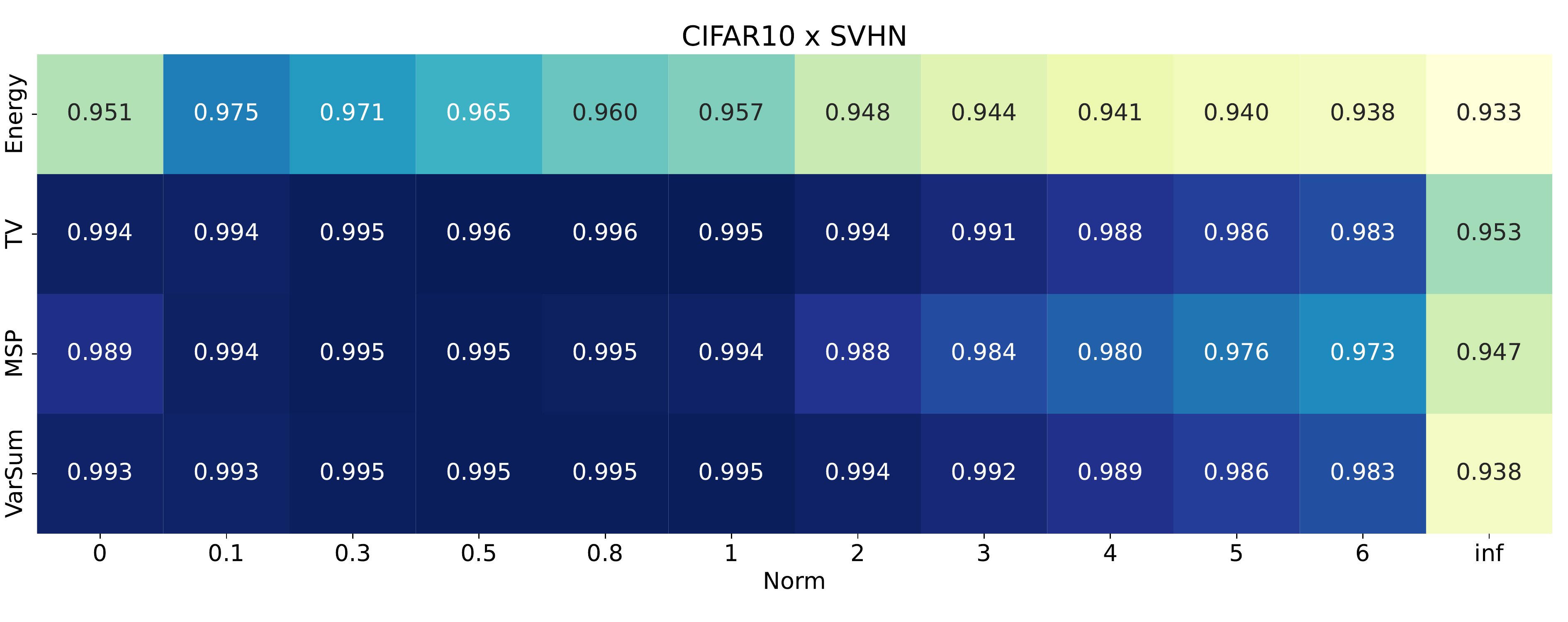}
    % \label{fig:my_label}
\end{figure}
\begin{figure}[h!]
    \centering
    \includegraphics[width=1\linewidth]{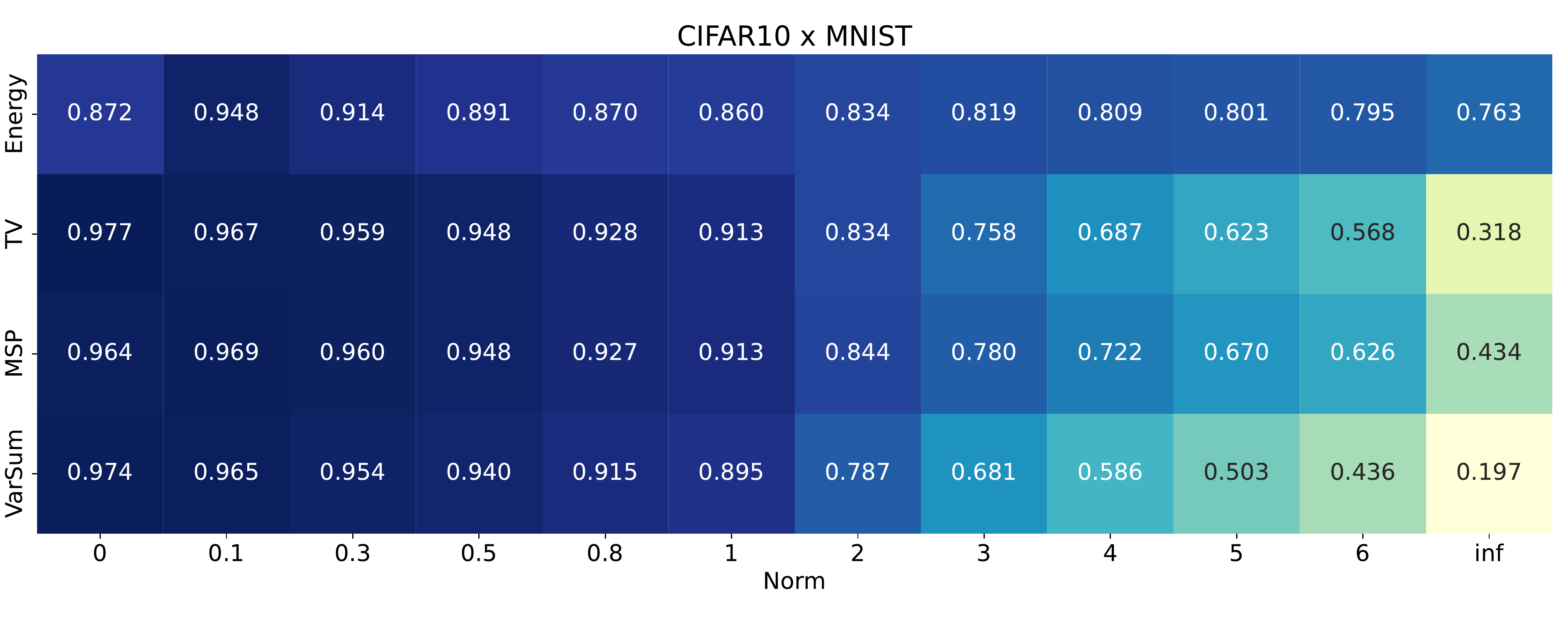}
    % \label{fig:my_label}
\end{figure}

%%%% ImageNet - 
\begin{figure}[h!]
    \centering
    \includegraphics[width=1\linewidth]{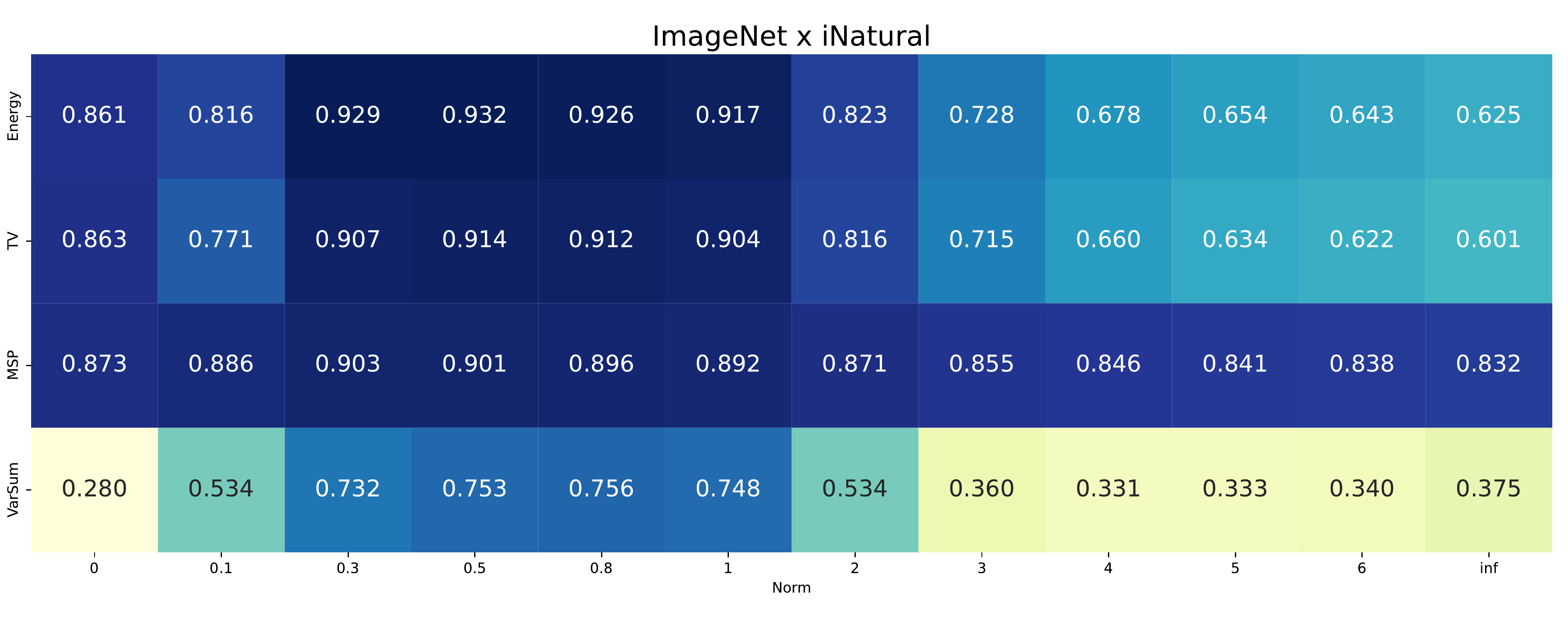}
    % \label{fig:my_label}
\end{figure}
\begin{figure}[h!]
    \centering
    \includegraphics[width=1\linewidth]{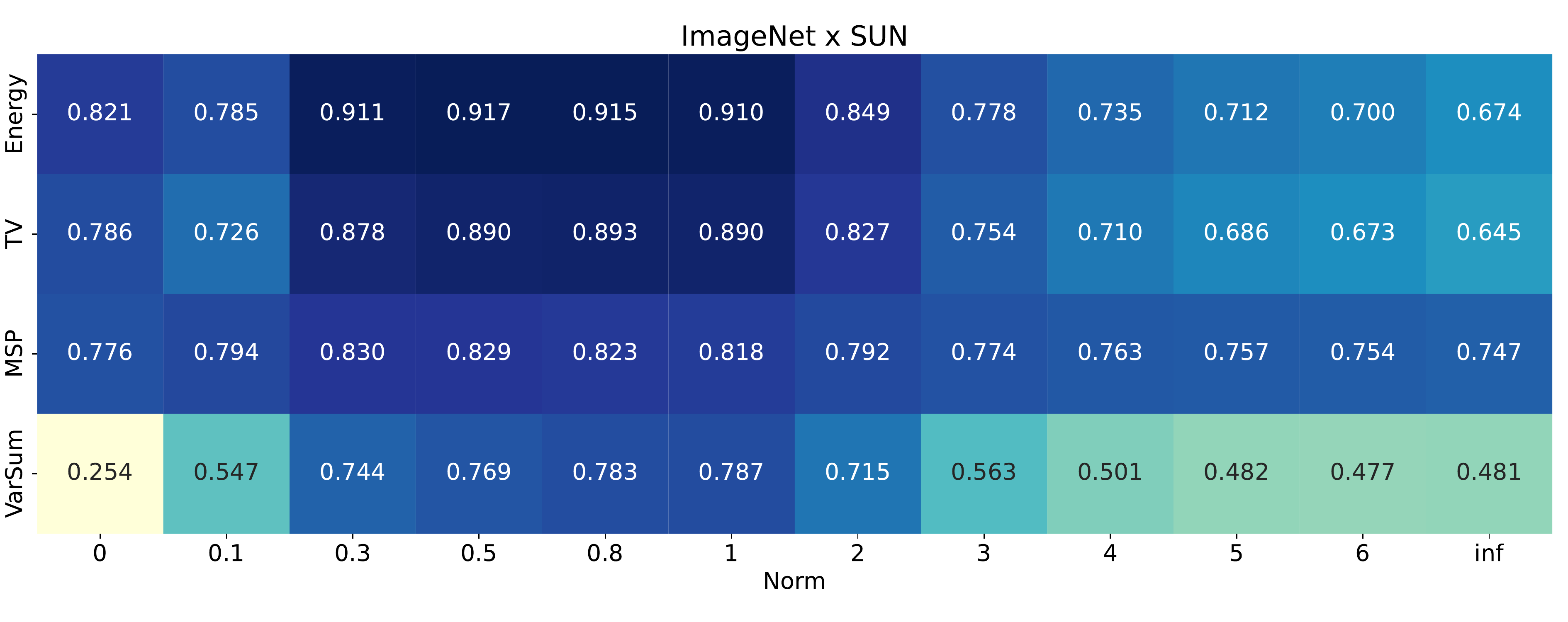}
    % \label{fig:my_label}
\end{figure}
\begin{figure}[h!]
    \centering
    \includegraphics[width=1\linewidth]{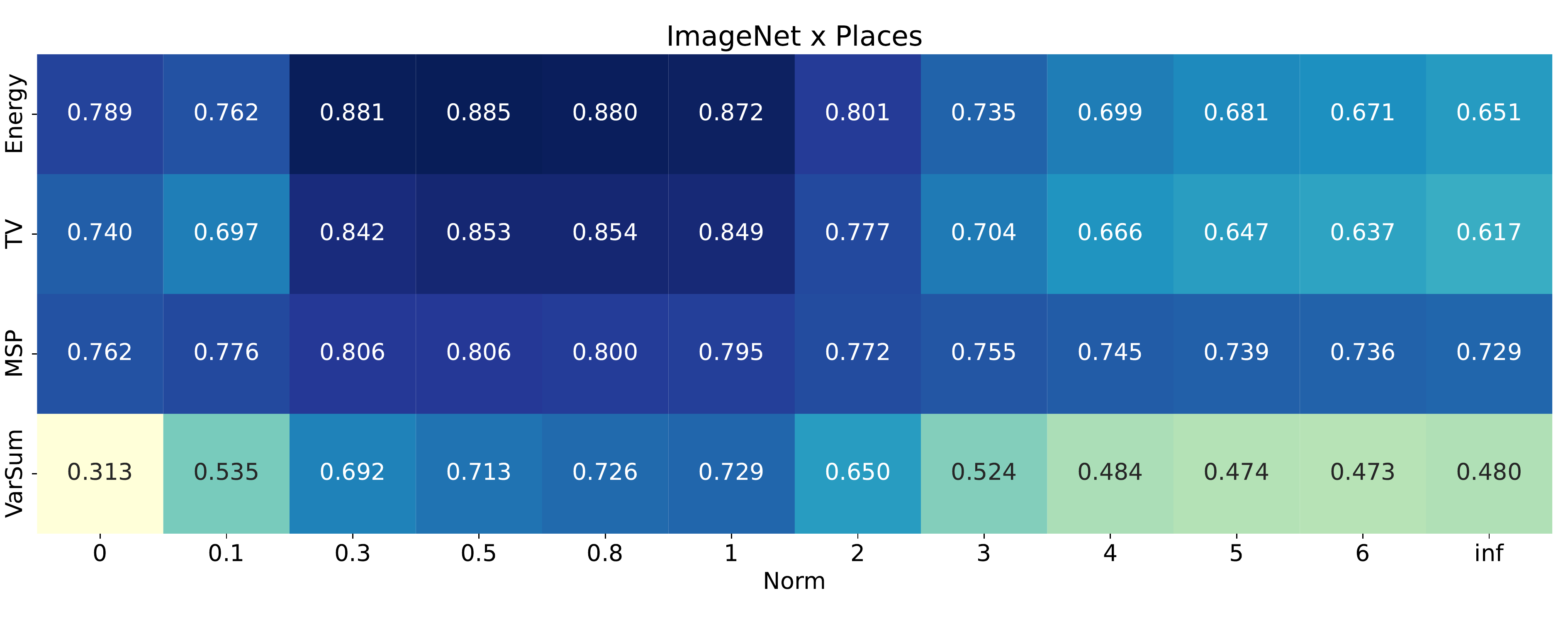}
    % \label{fig:my_label}
\end{figure}
\begin{figure}[h!]
    \centering
    \includegraphics[width=1\linewidth]{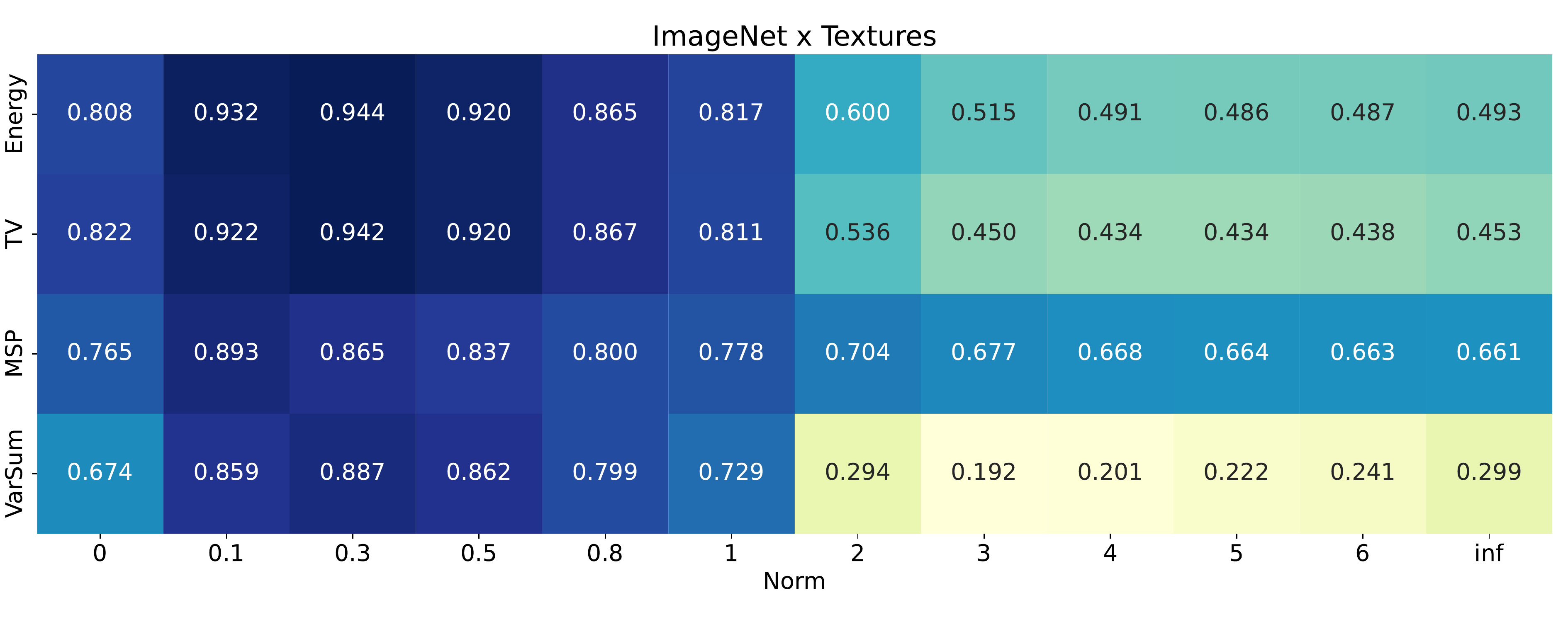}
    % \label{fig:my_label}
\end{figure}

\end{document}